\documentclass[twoside]{article}

\usepackage[accepted]{aistats2026} 
\usepackage[round]{natbib}

\usepackage{hyperref}       
\hypersetup{
    colorlinks,
    linkcolor={blue!50!black},
    citecolor={blue!50!black},
    urlcolor={blue!50!black}
}
\usepackage{url}            
\usepackage{booktabs}       
\usepackage{amsfonts}       
\usepackage{nicefrac}       
\usepackage{xcolor}         
\usepackage{amsmath}
\usepackage{cleveref}
\usepackage{bm}
\usepackage{tikz}
\usetikzlibrary{bayesnet}
\usetikzlibrary{arrows}
\bibpunct[, ]{(}{)}{,}{a}{}{,}%
\usepackage{color}
\usepackage{graphicx}
\usepackage{caption}
\usepackage{algorithm}
\usepackage{algpseudocode}

\newtheorem{proposition}{Proposition}
\DeclareMathOperator{\lnq}{\ln q}
\usepackage{subfig}
\usepackage[titletoc]{appendix}
\usepackage[dvipsnames]{xcolor}

\definecolor{myblue}{RGB}{68, 119, 170}
\definecolor{myorange}{RGB}{204, 187, 68}
\definecolor{mygreen}{RGB}{34, 136, 51}
\definecolor{myred}{RGB}{238, 102, 119}

\begin{document}
\newcommand{\JL}[1]{\textcolor{orange}{\textsf{#1}}}
\newcommand{\ZB}[1]{\textcolor{blue}{\textsf{#1}}}
\newcommand{\TODO}[1]{\textcolor{green}{\textsf{#1}}}
\newcommand{\placeholder}[1]{\textcolor{red}{\textsf{#1}}}
\newcommand{\dd}{\ \textrm{d}}

%
\runningtitle{Incorporating Expert Knowledge into Bayesian Causal Discovery of Mixtures of DAGs}

%

\twocolumn[

\aistatstitle{Incorporating Expert Knowledge into Bayesian Causal Discovery of Mixtures of Directed Acyclic Graphs}

\aistatsauthor{Zachris Björkman \And Jorge Loría \And  Sophie Wharrie \And Samuel Kaski}

\aistatsaddress{Aalto University \And  Aalto University \And University of Melbourne \And Aalto University\\Manchester University\\Ellis Institute Finland} 

]

\begin{abstract}
  Bayesian causal discovery benefits from prior information elicited from domain experts, and in heterogeneous domains any prior knowledge would be badly needed. However, so far prior elicitation approaches have assumed a single causal graph and hence are not suited to heterogeneous domains. We propose a causal elicitation strategy for heterogeneous settings, based on Bayesian experimental design (BED) principles, and a \textit{variational mixture structure learning} (VaMSL) method---extending the earlier \textit{differentiable Bayesian structure learning} (DiBS) method---to iteratively infer mixtures of causal Bayesian networks (CBNs). We construct an informative graph prior incorporating elicited expert feedback in the inference of mixtures of CBNs. Our proposed method successfully produces a set of alternative causal models (mixture components or clusters), and achieves an improved structure learning performance on heterogeneous synthetic data when informed by a simulated expert. Finally, we demonstrate that our approach is capable of capturing complex distributions in a breast cancer database.
\end{abstract}

\section{INTRODUCTION} 
\label{sec:introduction}
Prior elicitation has been called ``a central part of a Bayesian workflow'' \citep{mikkolaPriorKnowledgeElicitation2024} and is especially important in settings where there are few observations but practitioners have considerable expertise. Domains of low data/high expertise are common in causal modeling which has recently been stressed in the machine learning field \citep{kitsonSurveyBayesianNetwork2023}. Indeed, the need to involve domain experts when determining and validating causal models has been stressed, as the modeling may involve strong assumptions that are difficult to falsify \citep{hasanSurveyCausalDiscovery2023,brouillardLandscapeCausalDiscovery2024}. Others have listed further benefits, including reducing noise \citep{kitsonSurveyBayesianNetwork2023} and improved inference \citep{eggelingStructurePriorsLearning2019}.
\begin{figure}[t!]
    \centering
        \subfloat{\begin{tikzpicture}[scale=0.2\linewidth]
    \node[latent] (alpha1) {$\alpha_1$};%
    \node[latent,right=of alpha1,xshift=-0.25cm] (beta1) {$\beta_1$}; %
    \node[latent,below=of alpha1] (gamma1) {$\gamma_1$}; %
    \node[latent,right=of gamma1,xshift=-0.25cm] (delta1) {$\delta_1$}; %
    \edge [color=myblue, very thick] {alpha1} {gamma1, delta1}
    \edge [color=myblue, very thick]  {beta1} {alpha1}
    \edge [color=myblue, very thick]  {gamma1} {delta1}

    \node[latent,below=of alpha1,yshift=-1cm] (alpha2) {$\alpha_2$};%
    \node[latent,right=of alpha2,xshift=-0.25cm] (beta2) {$\beta_2$}; %
    \node[latent,below=of alpha2] (gamma2) {$\gamma_2$}; %
    \node[latent,right=of gamma2,xshift=-0.25cm] (delta2) {$\delta_2$}; %
    \edge [color=myorange, very thick] {beta2} {gamma2}
    \edge [color=myorange, very thick] {gamma2} {alpha2}
    \edge [color=myorange, very thick] {delta2} {alpha2, gamma2}
    
     \plate [inner sep=.2cm,yshift=.1cm] {plate2} {(alpha1) (alpha2) (beta1) (beta2) (gamma1) (gamma2) (delta1) (delta2)} {}; %

     \node[draw=none, above=of alpha1, xshift=0.70cm, yshift=-0.5cm] {Ground truth};
\end{tikzpicture}}
        \subfloat{\begin{tikzpicture}[scale=0.2\linewidth]
    \node[latent] (alpha1) {$\alpha_1$};%
    \node[latent,right=of alpha1,xshift=-0.25cm] (beta1) {$\beta_1$}; %
    \node[latent,below=of alpha1] (gamma1) {$\gamma_1$}; %
    \node[latent,right=of gamma1,xshift=-0.25cm] (delta1) {$\delta_1$}; %
    \edge [color=myblue, very thick] {alpha1} {gamma1}
    \edge [color=myblue, very thick]  {beta1} {alpha1}
    \edge [color=myblue, very thick]  {gamma1} {delta1}
    \edge [color=mygreen, very thick] {alpha1} {delta1}

    \node[latent,below=of alpha1,yshift=-1cm] (alpha2) {$\alpha_2$};%
    \node[latent,right=of alpha2,xshift=-0.25cm] (beta2) {$\beta_2$}; %
    \node[latent,below=of alpha2] (gamma2) {$\gamma_2$}; %
    \node[latent,right=of gamma2,xshift=-0.25cm] (delta2) {$\delta_2$}; %
    \edge [color=myorange, very thick] {gamma2} {alpha2}
    \edge [color=myorange, very thick] {delta2} {alpha2, gamma2}
    \edge [color=mygreen, very thick] {beta2} {gamma2}
    
     \plate [inner sep=.2cm,yshift=.1cm] {plate2} {(alpha1) (alpha2) (beta1) (beta2) (gamma1) (gamma2) (delta1) (delta2)} {}; %

     \node[draw=none, above=of alpha1, xshift=0.65cm, yshift=-0.5cm] {Our method};
\end{tikzpicture}}
        \raisebox{10mm}{
            \subfloat{\begin{tikzpicture}[scale=0.2\linewidth]
    \node[latent] (alpha) {$\alpha$};%
    \node[latent,right=of alpha,xshift=-0.25cm] (beta) {$\beta$}; %
    \node[latent,below=of alpha] (gamma) {$\gamma$}; %
    \node[latent,right=of gamma,xshift=-0.25cm] (delta) {$\delta$}; %
    \edge [color=myblue, very thick] {alpha} {delta}
    \edge [color=myblue, very thick] {beta} {alpha}
    \edge [color=myorange, very thick] {beta} {gamma}
    \edge [color=myred, very thick] {beta} {delta}
    \edge [color=myorange, very thick] {gamma} {alpha}
    
     \plate [inner sep=.2cm,yshift=.1cm] {plate2} {(alpha) (beta) (gamma) (delta)} {}; %

     \node[draw=none, above=of alpha1, xshift=0.65cm, yshift=-0.5cm] {DiBS};

\end{tikzpicture}}}
    \caption{Example of inference using a mixture model as compared to a non-mixture model with ground truth graphs with two components (\textit{left}), maximum-a-posteriori (MAP) graphs from our proposed mixture method (\textit{center}) with elicited edges (in \textcolor{mygreen}{\textbf{green}}), and MAP graph of non-mixture method DiBS (\textit{right}) without edge elicitation. Edges from the first graph are shown in \textcolor{myblue}{\textbf{blue}}, and edges from the second graph are in \textcolor{myorange}{\textbf{yellow}}. Incorrectly inferred edges are shown in \textcolor{myred}{\textbf{red}}.
    As shown, our method recovers the separate graphs while the non-mixture method finds one graph that tries to satisfy the contradictory information that the heterogeneous data encodes. For details see Appendix~\ref{app:num-settings}.}
    \label{fig:VaMSL_intro_example}
\end{figure}

When correctly specified, causal models remain valid even under \textit{interventions}, changes in the system that alter the joint distribution---which have been used to conceptualize several problems faced by the machine learning field \citep[e.g., distribution shifts;][]{scholkopfCausalAnticausalLearning2012, budhathokiWhyDidDistribution2021}. To infer such models, causal discovery methods aim to encode causal relationships as causal Bayesian networks (CBN) or structural causal models \citep{vowelsDyaDAGsSurvey2022}. Additionally, many have proposed Bayesian inference \citep[e.g.,][]{lorchDiBSDifferentiableBayesian2021a, fariaDifferentiableCausalDiscovery2022, heckermanTutorialLearningBayesian2022} to yield a posterior over probable causal models, addressing various modeling uncertainties, and offering a formal way to incorporate background knowledge through the prior. Even so, to date, no priors over graph structure have reached ``universal acclaim'' \citep{eggelingStructurePriorsLearning2019}. To leverage any information an expert might have about the causal relationships in the graph, we therefore propose both a novel informative graph prior to incorporate expert beliefs in the form of \textit{hard} and \textit{soft constraints} \citep{kitsonSurveyBayesianNetwork2023} and a querying strategy based on Bayesian experimental design (BED) principles \citep{rainforthModernBayesianExperimental2024} to elicit causal information from an expert.

However, even with an informative prior, the assumption of a single causal graph may be too restrictive to encode an expert's beliefs. Recent surveys \citep{brouillardLandscapeCausalDiscovery2024, markhamDistanceCovariancebasedKernel2022} in domains such as biology and medicine have pointed out that the common assumption of a \textit{single} underlying causal graph may not hold. Simultaneously, the promising alternative of mixtures of causal graphs needs more study \citep{variciSeparabilityAnalysisCausal2024}. To address the ``heterogeneous'' \citep{glymourReviewCausalDiscovery2019} setting in a Bayesian way, we propose a \textit{variational mixture structure learning} (VaMSL)\footnote{Source code in Python JAX \citep{jax2018github} available at \href{https://github.com/ZacBjo/VaMSL_proj}{https://github.com/ZacBjo/VaMSL\_proj}.} method to infer mixtures of CBNs. Our method extends the inference of the differentiable Bayesian structure learning (DiBS) method \citep{lorchDiBSDifferentiableBayesian2021a} from single CBNs to mixtures, enabling the specification of priors based on expert beliefs on heterogeneity. Although mixtures of causal graphs have been gaining interest \citep{saeedCausalStructureDiscovery2020,stroblCausalDiscoveryMixture2023,variciSeparabilityAnalysisCausal2024} and \citet{rittel_expressiveness_2025} also conclude the increased expressivity of mixtures can be beneficial for Bayesian causal discovery, few methods carry out Bayesian inference in the joint space of graphs structures and model parameters. These consist of works by \citet{castellettiBayesianLearningMultiple2020} and \citet{marchantCovariateDependentMixture2025}; however, their methods are limited to linear BNs and neither incorporates expert knowledge, making our method the first to carry out joint inference on both linear and non-linear BNs while incorporating expert knowledge. As is common, we assume causal faithfulness and causal sufficiency \citep{brouillardLandscapeCausalDiscovery2024}, within each component.

By combining VaMSL with the queried causal relationships, our method can learn mixtures of either linear or nonlinear CBNs in heterogeneous settings that arise in real-world scenarios. Figure~\ref{fig:VaMSL_intro_example} demonstrates this with a toy example using a two-component mixture with four nodes ($\alpha,\beta,\gamma,\delta$). Further, the proposed mixture method incorporates expert knowledge of the data generating process (DGP) to aid in identifying the graphs. Our results show that our proposed method can identify homogeneous subsets in heterogeneous data with improved structure learning. To summarize, our contributions are the following:
\begin{itemize}
    \item A novel informative (latent) graph prior to incorporate expert knowledge into causal discovery.
    \item A BED principled elicitation strategy to select optimal queries.
    \item A variational method for inference of mixtures of linear BNs \textbf{and}, for the first time, mixtures of nonlinear BNs.
\end{itemize}

\paragraph{Notation and Context of our Contributions within the Literature}
\label{sec:background}
Consider a directed acyclic graph (DAG) $G$ and parameters $\Theta$. The tuple $(G,\Theta)$ encodes a Bayesian network (BN), when the joint distribution factorizes according to the directed relationships in $G$, that is: $p(\mathbf{x}\mid G, \Theta) = \prod_i p(x_i\mid \textrm{pa}_G(x_i), \Theta)$, where $\textrm{pa}_G(x_i)$ corresponds to the nodes that have a directed edge towards $x_i$---the child. 
Causal BNs are BNs where the edges are \textit{causal} relationships, meaning that each child ($x_i$) is an \textit{effect} for which each parent is a \textit{direct cause}. With Bayesian inference we want to leverage observations $\mathcal{D}=\{\mathbf{x}_n\}_{n=1}^N$, where $\mathbf{x}_n\in\mathbb{R}^d$, to infer posteriors used to compute expectations for graph features of the form $p(f\mid \mathcal{D}) = \sum_G p(G\mid \mathcal{D})f(G)$, as described by \citet{friedmanBeingBayesianNetwork2000}; although in this work we focus on the more general case of inferring both graphs and parameters as $p(G,\Theta\mid \mathcal{D})$. Inference in the graph space is challenging, since it is a discrete space. To this end, \citet{lorchDiBSDifferentiableBayesian2021a} proposed 
DiBS, building on the work by \citet{zhengDagsNoTears2018}, which enables the use of gradient optimization methods for Bayesian inference in the (discrete) graph space via a latent continuous graph embedding space $\mathbf{Z} = [\mathbf{U},\mathbf{V}]$, with $\mathbf{U},\mathbf{V}\in \mathbb{R}^{\ell \times d}$, where $\ell$ is a latent embedding dimensionality and $d$ is the number of variables in the BN. A sample from $\mathbf{Z}$ parametrizes a temperature-dependent distribution over graphs according to $p_\omega(G\mid \mathbf{Z}) = \prod_{i\not=j}\textrm{Ber}(G_{ij}\mid G_{\omega}(\mathbf{Z})_{ij})$ with ${G_{\omega}(\mathbf{Z})_{ij} = \sigma_\omega (\mathbf{U}_{\cdot i} \cdot \mathbf{V}_{\cdot j})}$, where $\omega$ is an inverse temperature and $\sigma_\omega(\lambda) = 1/\left(1+\exp(-\omega \lambda)\right)$ is the logistic function. We refer to the matrix of edge probabilities encoded by $G(\mathbf{Z})$ as a \textit{soft graph} and suppress the dependence on $\omega$ when possible. The use of gradient methods in DAG search is enabled by incorporating a function $h$ that characterizes the cyclicity of a given graph $G$, such that $h(G)=0$ if and only if $G$ is acyclic. Additionally, to infer DAGs, DiBS uses another temperature parameter $\beta$ that is annealed over the optimization scheme, degenerating the prior probability of generating a cyclic graph, such that $\left.p_\beta(\mathbf{Z}')\right|_{h(G') \not=0,\ G'\sim p(G\mid \mathbf{Z}')}$ goes to zero when $\beta\to\infty$. For more details see \citet{lorchDiBSDifferentiableBayesian2021a}. 

A standard (finite) mixture model consists of distributions over a simplex of mixing weights $\bm{\pi} = \{\pi_k\}_{k=1}^K$, \textit{one-hot} assignment variables $\bm{c}_n = \{c_{nk}\}_{k=1}^K,{n=1,\ldots,N}$, and a \textit{mixture likelihood} for the observations. In the case of a mixture of BNs, consisting of $K$ tuples of graphs and parameters denoted $\{(G_k,\Theta_k)\}_{k=1}^K$, these distributions are:
\begin{align}\label{eq:mixture}
    &\bm{\pi} \sim \textrm{Dir}(\bm{\alpha}),
    \;\; \bm{c}_n \sim \textrm{Cat}(\bm{\pi}), \textrm{ and } \nonumber \\ 
    &p(\mathbf{x}_n \mid G, \Theta, \bm{c}_n) = \prod_k\nolimits p(\mathbf{x}_n \mid G_k, \Theta_k)^{c_{nk}},
\end{align}
respectively. A common technique for inference in these models is through variational approximations. An optimization scheme is then used to maximize the \textit{evidence lower bound} (ELBO), which corresponds to minimizing the Kullback-Leibler divergence between the variational distribution and the posterior. As the standard approach of DiBS, we use Stein variational gradient descent \citep[see,][]{liuSteinVariationalGradient2016} in each component; for a review of variational methods we refer the reader to \citet{zhangAdvancesVariationalInference2018a}. 

Inferring the true causal DGP from covariates in an observational setting is a challenging problem. Indeed, the alternative of compensating for a lack of data by making use of expert knowledge in causal discovery (structure learning) methods \citep[e.g.,][]{lorchDiBSDifferentiableBayesian2021a} has long been acknowledged in the field \citep{canoMethodIntegratingExpert2011}. Early works combining causal discovery from observations with expert information to infer directed graphical models include \citet{buntineTheoryRefinementBayesian1991} and \citet{heckermanLearningBayesianNetworks1995}. Here we consider structured informative priors. These priors are usually distinguished by whether they consist of \textit{hard} or \textit{soft} constraints \citep{angelopoulosBayesianLearningBayesian2008, constantinouImpactPriorKnowledge2023}. Hard constraints require (forbid) given edges, and soft constraints merely bias the search to including (excluding) them. \citet{angelopoulosBayesianLearningBayesian2008} review Bayesian methods incorporating informative priors on structure, noting that ``\textit{most BN learning approaches make no significant attempt to integrate prior knowledge into the learning process}'' and that most priors are chosen for ``pragmatic reasons''. Based on a lack of ``systematic comparisons'', \citet{eggelingStructurePriorsLearning2019} conduct a general investigation into common structure priors, favoring sparsity over the standard uniform choice. They support their arguments both analytically and experimentally by comparing multiple common structure priors.  \citet{constantinouImpactPriorKnowledge2023} review a variety of methods to incorporate prior knowledge in causal discovery. Notably, they do not include soft constraints based on elicited probabilities. Further causal discovery methods incorporating hard constraints include \citet{wang_prior-knowledge-driven_2020}, \citet{hasan_kcrl_2022}, \citet{hasan_kgs_2023}, and \citet{sun_nts-notears_2023}. However, our method incorporates \textit{both} hard and soft constraints as the latter are less strict.  \citet{rittelSpecifyingPriorBeliefs2023} propose several priors to use on inference of a causal graph (including a prior for DiBS) which can themselves be expert-informed. However,  they do not consider any possibility of actively learning the graph, and their implementations are not publicly available. Most recently, \citet{marchantCovariateDependentMixture2025} propose a model (with an accompanying inference method) for mixtures of BNs, where the mixture probabilities depend on covariates, though the possibility of expert-informed priors is only addressed by indicating which of the covariates are ``modifiable''.

To optimally select edges to elicit from an expert, we use a BED approach \citep{rainforthModernBayesianExperimental2024}. Previously, BED has been applied to causal discovery to pick maximally informative intervention experiments to run \citep[see,][]{tothActiveBayesianCausal2022,zhang_goal-oriented_2025, zemplenyi_bayesian_2023, tigas_differentiable_2023}, our goal is instead prior elicitation. While this is related to the work by \citet{ibrahimTargetedCausalElicitation2022}, they do not try to infer full causal graphs instead focusing on inferring necessary interventions to regulate a target variable. 

Applying BED  requires a model; in our case of expert elicitation, a model of the expert. As such, our model includes two parts: the currently inferred latent graph posterior $p(\mathbf{Z}\mid\mathcal{D})$ and a simulator of responses from the expert. The former is a probability distribution over parameters for the latter. We incorporate the simulator by using the \textit{expected information gain} (EIG) \citep{bernardoExpectedInformationExpected1979, lindleyMeasureInformationProvided1956} as a utility function, that we maximize over the set of possible queries and prompt the selected query to the expert. 
Maximizing the EIG corresponds to obtaining the largest (Shannon) information between the prior and (potential) posterior distributions, thereby optimally leveraging the expert.
  

The rest of the article is organized as follows. In Section~\ref{sec:elicitation} we present and explain the importance of elicitation in causal settings and our proposed priors. Section~\ref{sec:Method} exemplifies our methodological contributions in the learning of a mixture of structured graphs. We demonstrate the performance of our approach in Section~\ref{sec:results}, by simulations and an application to a cancer identification problem in the UCI machine learning repository. We finalize with pointing out conclusions and future directions in Section~\ref{sec:conclusions}.


\section{ELICITATION OF CAUSAL-EDGE INFORMATION} \label{sec:elicitation} 
For elicitation in the causal setting we assume there is an underlying CBN which corresponds to the DGP. Additionally, we assume that the queried expert has, possibly uncertain, knowledge of the underlying CBN (otherwise, why ask them?). We will query the expert on their beliefs regarding the edges in the graph of the CBN; we denote these edge-specific beliefs by $p(\psi_{ij}\mid \mathcal{K}_{ij})$, for each edge $i\rightarrow j$, where $\mathcal{K}_{ij}$ is the knowledge the expert is basing their beliefs on. 
Further, we assume the distribution which gives rise to their beliefs factorizes over the edges of the graph. The joint distribution of the edge beliefs is then 
\begin{equation*}
    p(\Psi\mid \mathcal{K}) = \prod_{i\not=j} p(\psi_{ij}\mid \mathcal{K}_{ij}).
\end{equation*}
The difficulty of eliciting beliefs as a probability distribution is well-documented \citep{ohaganUncertainJudgementsEliciting2006, mikkolaPriorKnowledgeElicitation2024}; we therefore only elicit a probability which we interpret as the mode of the distribution of edge-beliefs---denoted by $\psi_{ij}^* = \textrm{argmax}_{\psi_{ij}\in[0,1]}\ p(\psi_{ij}\mid \mathcal{K}_{ij})$. 
We distinguish between hard and soft constraints, and denote the hard constraint responses by $\mathcal{D}_H = \{\psi_{ij}^*\in \{0,1\} : \text{ the edge $i\rightarrow j$ has been elicited}\}$ and the soft constraint responses as $\mathcal{D}_S = \{\psi_{ij}^*\in (0,1) : \text{ the edge $i\rightarrow j$ has been elicited}\}$. 
We do \textbf{not} claim to observe the random variable $\psi_{ij}$ nor its distribution $p(\psi_{ij}\mid \mathcal{K}_{ij})$---in general we cannot observe either. Rather we interpret the response we obtain as the mode of the distribution $p(\psi_{ij}\mid \mathcal{K}_{ij})$. This approach can be extended to other summary statistics than the mode, but that is left for future work.

Our aim is to incorporate the expert's responses ($\mathcal{D}_H, \mathcal{D}_S$) into the inference of the data generating CBN through an informative graph \emph{prior}.
We want to minimize the number of queries since the act of querying is time-consuming and potentially expensive.  
Moreover, an exhaustive query of all the edges is infeasible even for graphs with a moderate number of nodes. As mentioned by \citet{rittelSpecifyingPriorBeliefs2023}, the potential issue of an expert expressing cyclic beliefs (forbidden by the DAG assumption), is mitigated by the cyclicity regularizer that the DiBS method leverages as part of its prior. The cyclicity regularizer assures the inferred graphs are DAGs and allows us define to our querying strategy in terms of edge beliefs, rather than orderings \citep[e.g.,][]{buntineTheoryRefinementBayesian1991} or other DAG-ensuring constructs. This significantly reduces the burden of the expert while the interpretability of the query can contribute to the quality of the responses. We address the need to minimize queries by relying on BED principles to select the most informative edges for our prior.
\subsection{Selecting Elicitation Queries by Bayesian Experimental Design}\label{sec:elicitation_BED}
We elicit the expert responses $(\mathcal{D}_H, \mathcal{D}_S)$ that are used to construct the elicitation prior in the following way.
Suppose we have access to a posterior distribution over latent graph embeddings $p(\mathbf{Z} \mid \mathcal{D})$, for instance using DiBS, where each $\mathbf{Z}$ encodes probabilities $G(\mathbf{Z})_{ij} \in [0, 1]$ for each edge $i \rightarrow j$ in a causal graph. Consider a query of a specified edge as an experiment $\xi_{ij} \in \Xi$, which results in a response $\psi_{ij}^* \in [0, 1]$. This response is the expert's belief (probability) for the edge existing in the causal graph $G$. Depending on the application, the design space $\Xi$ can vary. However, we will assume it consists of all the edges in the causal graph (barring self-loops). For instance, each observation could contain variables from two snapshots at times $t_1<t_2$, logically we would not want any variable at time $t_2$ to have an edge directed towards variables at time $t_1$; such possibilities can be excluded from $\Xi$.

The BED framework \citep{rainforthModernBayesianExperimental2024} requires a parameter prior as well as a simulator for experiment outcomes. In our case these will be $p(\mathbf{Z} \mid \mathcal{D})$ and $p(\psi_{ij}^*\mid \mathbf{Z}, \xi_{ij})$, respectively. To simulate querying the expert about their edge belief, we propose the simulator
\begin{align*}
    &p(\psi_{ij}^* \mid \mathbf{Z}, \xi_{ij})\\ &= \textrm{Beta}(\alpha_s {G(\mathbf{Z})_{ij}} +1, \beta_s{(1-G(\mathbf{Z})_{ij}})+1),\notag
\end{align*}
where setting $\alpha_s = \beta_s$ yields a distribution with a mode of $G(\mathbf{Z})_{ij}$ and the variance of the simulator is fully determined by the choice of value for $\alpha_s$. In our experiments we set $\alpha_s = \beta_s=10$. 

We propose to use the EIG to evaluate and select queries to pose to the expert. The EIG serves as a utility function, and we can simulate the possible future responses from the expert given by the current model $p(\mathbf{Z} \mid \mathcal{D})$. Maximizing the EIG finds the edge that is expected to provide the most information for posterior inference; this enables an efficient convergence to the causal graph of interest. Specifically, we pick the optimal query $\xi^{*}_{ij}$ that satisfies
\begin{align}\label{eq:BED_EIG_max} 
    &\underset{\xi_{ij} \in \Xi}{\textrm{argmax}}\hspace{1mm} \textrm{EIG}(\xi_{ij})  \\
    &= \underset{\xi_{ij} \in \Xi}{\textrm{argmax}}\hspace{1mm}                           
        \mathbb{E}_{p(\mathbf{Z}\mid \mathcal{D})p(\psi_{ij}^*\mid \mathbf{Z}, \xi_{ij})} 
        \left[\ln p(\psi_{ij}^* \mid \mathbf{Z}, \xi_{ij})\right.\notag \\ 
        &\quad\left.- \ln p(\psi_{ij}^* \mid \xi_{ij})\right].\notag 
\end{align}
For this we use a nested Monte Carlo approximation $\hat{\mu}_{\textrm{NMC}}$ (see Appendix~\ref{app:approximating_EIG} for details). 
This requires sampling $\mathbf{Z}^{(n)} \sim p(\mathbf{Z} \mid \mathcal{D})$ from the parameter prior. This is trivial to do from the particle posterior that the Stein variational gradient descent (SVGD) instantiation of DiBS provides. We note that, in case we are only interested in binary responses $\psi_{ij}^* \in \{0, 1\}$, we can use a Bernoulli distributed simulator and tractably compute the outer expectation of Eq.~\eqref{eq:BED_EIG_max} across ${\psi_{ij}^*}$, enabling the use of the Rao-Blackwellized EIG estimator $\hat{\mu}_{\textrm{RB}}$. Full details of how these Monte Carlo approximations are computed, and alternatives for EIG approximations, are described by \citet{rainforthModernBayesianExperimental2024}.
\subsection{Elicited Informative Graph Prior}
\label{sec:informative_elicitation_graph_prior}
Our goal is to infer CBNs conditional on elicited responses ($\mathcal{D}_H,\mathcal{D}_S$) from the expert in the form of hard and soft constraints, respectively, that were selected as described above. \textit{Hard constraints} directly shrink the considered graph space and have been called ``the most significant form of prior knowledge'' \citep{angelopoulosBayesianLearningBayesian2008}. Allowing experts to fully rule out edges that do not coincide with their fundamental understanding of the causal system under investigation (e.g.,~in biomedical applications a biomarker cannot ``cause'' the age of a patient) or require edges they deem necessary. Imposing hard constraints of absence (existence) requires $p(G) = 0$ whenever $G_{ij}=1$ ($G_{ij}=0$),  which is achieved by setting the prior of such graphs to satisfy this condition. Notably, in DiBS this is done by $0/1$-masking the appropriate edges when converting latent graph embeddings $\mathbf{Z}$ to (soft) graphs $G$, restricting the graph search to only the space of graphs satisfying the hard constraints. To incorporate \textit{soft constraints} we construct an informative graph prior based on Bayesian prior elicitation \citep{mikkolaPriorKnowledgeElicitation2024}.

To impose an informative prior on the latent space $\mathbf{Z}$, we propose a user model for the expert inspired by the \textit{imaginary observations} device \citep{consonniPriorDistributionsObjective2018, goodProbabilityWeighingEvidence1950a} and generate a hypothetical set of observations which the expert would condition on to form their opinion. 
We define a mapping $f: [0,1] \rightarrow \mathbb{N}^2$ from expert responses $\psi_{ij}^*$ to the tuple $\mathcal{K}_{ij}=(n_{ij}, k_{ij})$, representing a number of \textit{imagined observations} (trials) $n_{ij}$ and a number of successes $k_{ij}$ (observations of the existence of the causal edge $i\to j$). These imagined observations are part of a data generating story for an idealized experiment the expert has conducted to investigate the existence (absence) of a causal edge to form their posterior edge beliefs, such that for all $i\neq j$, $p(\psi_{ij}\mid \mathcal{K}_{ij}) \propto p(\mathcal{K}_{ij}\mid \psi_{ij})p(\psi_{ij})$. 

\begin{figure}[!htb]
    \begin{center}
    \begin{tikzpicture}
     \node[latent] (psi) {$\bm{\psi}_{ij}$};%
     \node[obs, right=of psi, xshift=0.3cm] (trials) {$\mathcal{K}_{ij}$};%
     \node[const, left=of psi,xshift=-0.3cm] (alphabeta) {$\{\alpha_0, \beta_0\}$};
     
     \edge {psi} {trials}
     \edge {alphabeta} {psi} 
    
     \plate [inner sep=.2cm,xshift=-.1cm] {plate1} {(psi) (trials)} {$ij$}; %
\end{tikzpicture}
    \caption{Generative model for expert edge beliefs ($\psi_{ij}$), with hyperparameters $\alpha_0,\beta_0$, and trials observed by the expert $(\mathcal{K}_{ij})$.}
    \label{fig:imaginary_observations_generative_model_main}
    \end{center}
\end{figure}
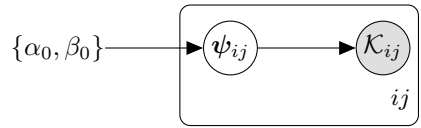
The generative model for the expert's beliefs follows the diagram shown in \Cref{fig:imaginary_observations_generative_model_main}. We assign beta priors for $i\neq j,\ p(\psi_{ij})=\textrm{Beta}(\psi_{ij}\mid \alpha_0, \beta_0)$ to represent the beliefs of the expert before they observed any causal edges (letting each edge prior share parameters). Additionally, we model the update of the expert's beliefs about an edge's existence as a beta-binomial conjugate distribution. For simplicity, we assume the idealized experiment the expert uses to verify a causal edge's existence is deterministic. Hence, if $i$ is a direct cause of $j$ (i.e., $i\to j$) then all trials confirm it (i.e., $k_{ij}=n_{ij}$), and vice versa (i.e., $k_{ij}=0$). Given that the queried responses $\psi_{ij}^* \in \mathcal{D}_{S}$ are the modes of the expert's posterior beliefs, conjugacy yields a unique mapping from an expert's prior $\textrm{Beta}(\alpha_0, \beta_0)$ to their posterior $p(\psi_{ij}\mid \mathcal{K}_{ij}) = \textrm{Beta}(\alpha_{ij}, \beta_{ij})$, with the (elicited) mode $\psi_{ij}^*=\frac{\alpha_{ij}-1}{\alpha_{ij} + \beta_{ij}-2}$. As the mode must have moved from the mode of the prior $\psi_{ij,0}^{*}=\frac{ \alpha_0-1}{ \alpha_0+ \beta_0-2}$ in the direction supported by observations, we derive that:
\begin{align}
    &f_{\alpha_0, \beta_0}(\psi_{ij}^*) \label{eq:mode_update}\\
    &=\begin{cases}
        \mathcal{K}_{ij} =  \left(n_{ij}=\lfloor\frac{\psi_{ij}^*(\alpha_0+\beta_0-2)-\alpha_0+1}{1-\psi_{ij}^*}\rfloor, k_{ij}=n_{ij}\right),\\
        \quad\textrm{if }\ \psi_{ij}^* > \psi_{ij,0}^{*};\\[4pt]
        \mathcal{K}_{ij} = \left(n_{ij} =\lfloor\frac{\alpha_0-1-\psi_{ij}^*(\alpha_0+\beta_0-2)}{\psi_{ij}^*}\rfloor, k_{ij}=0\right),\\
        \quad\textrm{if }\  \psi_{ij}^* < \psi_{ij,0}^{*}, \\
    \end{cases} \notag
\end{align}
where the flooring ensures that we only consider integer observations and the number of imaginary observations can be identified by rearranging the posterior mode. For more details, see Appendix~\ref{app:user_model}.

We denote the set of all mappings $\mathcal{D}_{\mathcal{K}}=f(\mathcal{D}_S)$ and construct the \textit{elicited informative graph prior} on the latent graph embeddings $\mathbf{Z}$:
\begin{align}
    p(\mathbf{Z} \mid \mathcal{D}_{S}) &\propto p( \mathbf{Z})p(f(\mathcal{D}_{S}) \mid \mathbf{Z})\\ 
    &= p(\mathbf{Z})\prod_{i\not=j}\nolimits p(\mathcal{K}_{ij} \mid G(\mathbf{Z})_{ij})\notag,
\end{align}
where the \textit{elicitation likelihood} \citep{mikkolaPriorKnowledgeElicitation2024} $p(\mathcal{K}_{ij} \mid G(\mathbf{Z})_{ij})$ is evaluated by parameterizing a binomial distribution with the soft graph encoded by $\mathbf{Z}$, with trials confirming (refuting) an edge viewed as $k$ successes ($n-k$ failures):
\begin{equation*}
    p(\mathcal{K}_{ij} \mid G(\mathbf{Z})_{ij}) = \textrm{Bin}(n=n_{ij},k= k_{ij} \mid p=G(\mathbf{Z})_{ij}),
\end{equation*}
where, in each case, both the binomial coefficient and one of the factors in the binomial probability mass function evaluate to 1 (as either $k_{ij}=n_{ij}$ or $k_{ij}=0$, see Eq.~\eqref{eq:mode_update}). Note that the informativeness of the prior is set by the choice of $\alpha_0, \beta_0$, as they determine how many imagined observations a given response from the expert maps to. Potential cyclic beliefs are not an issue, as the cyclicity regularizer in the DiBS prior $p(\mathbf{Z})$ collapses the latent graph space unto the DAG space during optimization, eliminating any possible cyclic beliefs in the posterior. For more details of our implementation, see Appendix~\ref{app:user_model_implementation}.
\subsection{Incorporating Expert Responses to Model Updates}
In our querying scheme, the expert is queried after a set number of optimization steps, using the currently inferred distribution as the parameter prior for the BED selected queries. Once the querying is completed, the structure learning method can continue, conditional on the new responses. We here assume the problem of label switching \citep{celeuxBayesianInferenceMixture1998} is addressed by the expert being able to indicate which component refers to which set of beliefs (e.g., by visual inspection of the current clustering). Moreover, expert information is a way to specify non-exchangeable priors, which prevents label switching \citep{betancourtIdentifyingBayesianMixture2017}. 

The number of times the optimization is interrupted to query the expert depends on how many times the expert can be queried as well as the total computational budget. Notably, in the case of our proposed model, described in \Cref{sec:Method}, the iterative coordinate-ascent variational inference (CAVI) procedure naturally enables the expert to be queried every time the clustering is updated and new information about identified subpopulations can be provided. Thus, the inference of our proposed mixture method, when paired with our elicited graph prior and used by a domain expert, is informed by responses to queries about edges they supply.
\section{(DIFFERENTIABLE) 
VARIATIONAL MIXTURE STRUCTURE LEARNING} 
\label{sec:Method}
Classically, a single graph has been assumed to be sufficient to explain the causal mechanisms of a real-world phenomena, however the possibility of several graphs explaining the data should not be ignored as the real world is often complex and cannot be easily described. Correspondingly, learning mixtures of models is more demanding than a single model, both computationally, and statistically. For graphs this difficulty compounds \citep{variciSeparabilityAnalysisCausal2024}. To alleviate this difficulty, we incorporate expert beliefs about the causal relationships into the inference procedure through the construction of component- and edge-specific informative graph priors.

To address data heterogeneity, we use a mixture of CBNs (Eq.~\eqref{eq:mixture}). We leverage the method proposed by \citet{lorchDiBSDifferentiableBayesian2021a} to infer a posterior with respect to CBNs with graphs in the (continuous) latent graph embedding space as well as the mixing weights and the assignment variables. This enables the use of gradient optimization to carry out the (otherwise discrete) search across the graph space. Our method applies variational inference to approximate the posterior as
\begin{equation*}
    q(\bm{\pi}, \mathbf{Z}, \Theta, C) \approx p(\bm{\pi}, \mathbf{Z}, \Theta, C\mid \mathcal{D}).
\end{equation*}
We refer to \citet{lorchDiBSDifferentiableBayesian2021a} for a detailed account of how inference over the latent variables $\mathbf{Z}_k$ corresponds to inference over graphs $G_k$ ($K=1$ in our setting). However, any inference algorithm that yields a posterior over CBNs can be used for VaMSL. As is common in VI, we make the mean-field assumption, approximating the posterior as the factorizing distribution:
\begin{align*} 
    q(\bm{\pi}, \mathbf{Z}, \Theta,C) 
    &= q(\bm{\pi})q(\mathbf{Z}, \Theta)q(C)\\
    &= q(\bm{\pi})\prod_k\nolimits q(\mathbf{Z}_k, \Theta_k)\prod_n\nolimits q(\bm{c}_{n})\notag.
\end{align*}
The factorizing distributions are iteratively optimized using CAVI. To optimize the variational CBN distributions $q(G, \Theta)$, we need to employ a Bayesian structure learning framework. In this work, we use VaMSL with the DiBS framework \citep{lorchDiBSDifferentiableBayesian2021a}, replacing inference over $G$ with the continuous relaxation $\mathbf{Z}$ and enabling the use of gradient-based optimization methods to carry out posterior inference. Before introducing the CAVI-steps employed to optimize the variational approximation of the posterior, we describe how we use the continuous relaxation over the graph space.
\subsection{(Differentiable) 
Generative Graph Mixture Model}\label{sec:generative_graph_mixture}
\begin{figure}[!ht]
    \begin{center}
    \scalebox{.75}{\begin{tikzpicture}
     \node[obs] (x) {$\bm{x}_n$};%
     \node[latent,left=of x,xshift=-0.25cm] (theta) {$\Theta_k$}; %
     \node[latent,below=of theta] (G) {$G_k$}; %
     \node[latent,left=of G,xshift=-0.25cm] (Z) {$\mathbf{Z}_k$}; %
     \node[latent,right=of x,xshift=0.25cm] (c) {$\bm{c}_n$}; %
     \node[latent,below=of c] (pi) {$\bm{\pi}$}; %
     \node[const, left=of pi,xshift=-0.25cm] (alpha) {$\bm{\alpha}$}; %
     \node[const, left=of Z,xshift=-1.75cm] (betaomega) {$\{\beta, \omega\}$}; %
     \node[obs, above=of Z] (hard_responses) {$\mathcal{D}_{H,k}$}; %
     \node[obs,left=of hard_responses,xshift=-0.25cm] (responses) {$\mathcal{D}_{S,k}$}; %
     
     \edge {theta, G, c} {x}
     \edge {G} {theta}
     \edge {Z} {G, responses}
     \edge {hard_responses} {G}
     \edge {pi} {c}
     \edge {alpha} {pi} 
     \edge {betaomega} {Z} 
    
     \plate [inner sep=.3cm,xshift=.02cm,yshift=.2cm] {plate1} {(x) (c)} {$N$}; %
     \plate [inner sep=.3cm,xshift=.02cm,yshift=.2cm] {plate2} {(G) (Z) (theta) (responses)} {$K$}; %
\end{tikzpicture}}
    \caption{Generative model for VaMSL using DiBS.}
    \label{fig:VaMSL_generative_model_main}
    \end{center}
\end{figure}
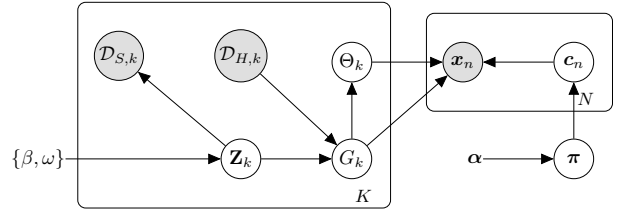

Figure~\ref{fig:VaMSL_generative_model_main} shows the DGP of our mixture model, using DiBS, and it corresponds to
\begin{align*}
    &p(\bm{\pi}, \mathbf{Z}, G, \Theta, C, \mathcal{D} \mid \mathcal{D}_S, \mathcal{D}_H)\\ 
    &= \ p(\bm{\pi})p(\mathbf{Z}\mid \mathcal{D}_S)p(G\mid \mathbf{Z}, \mathcal{D}_H)p(\Theta \mid G)\nonumber\\
    &\quad\times p(C\mid \bm{\pi})p(\mathcal{D}\mid C,G,\Theta) \nonumber\\
    &=\ p(\bm{\pi}) \prod_k\nolimits \left[ p(\mathbf{Z}_k\mid \mathcal{D}_{S,k})p(G_k\mid \mathbf{Z}_k, \mathcal{D}_{H,k})p(\Theta_k\mid G_k)\right]   \nonumber\\
    &\quad\times \prod_n\nolimits \left[ p(\bm{c}_n\mid \bm{\pi})\prod_k\nolimits p(\mathbf{x}_n\mid c_{nk}, G_k,\Theta_k)\right]\notag,
\end{align*}
where graphs with $d$ variables are represented by adjacency matrices $G_k \in \{0, 1\}^{d \times d}$. Each component of the graph embeddings $\mathbf{Z}_k = [\mathbf{U}^{(k)}, \mathbf{V}^{(k)}]$ with $\mathbf{U}^{(k)}, \mathbf{V}^{(k)}\in \mathbb{R}^{\ell \times d}$ takes the form described in Section~\ref{sec:background}. The parameter distribution over $\Theta$ is a modeling choice dependent on the chosen family of CBNs, where we consider both the linear and non-linear case. The data-likelihood corresponds to the mixture likelihood in Eq.~\eqref{eq:mixture}.

The generative graph prior of the latent graph embeddings $\mathbf{Z}_k$, encodes a distribution over directed graphs via the following generative model: 
\begin{align} \label{eq:generative_graph_prior}
    &p_\omega(G_k\mid \mathbf{Z}_k, \mathcal{D}_{H,k}) = \prod_{ij: \psi_{ij}^{*}\in \mathcal{D}_H} \psi_{ij}^{*} \\
    &\times \prod_{ij: \psi_{ij}^* \not \in \mathcal{D}_H} \textrm{Ber}\left(G_{k,(ij)}\mid \sigma_\omega (\mathbf{U}^{(k)}_{\cdot i} \cdot \mathbf{V}^{(k)}_{\cdot j})\right)  \notag,
\end{align}
where $G_{k,(ij)}$ refers to the edge $i \rightarrow j$ in component $k$. Recall that the \textit{hard} constraints $\psi_{ij}^*\in \mathcal{D}_H$ can only take values of zero or one, meaning that they rule out any graphs that violate them. Additionally, \citet{lorchDiBSDifferentiableBayesian2021a} define the \textit{soft graph}, the matrix of edge probabilities encoded by the latent graph $\mathbf{Z}_k$, as: $G_\omega(\mathbf{Z}_k)_{ij} = \sigma_\omega (\mathbf{U}^{(k)}_{\cdot i} \cdot \mathbf{V}^{(k)}_{\cdot j})$. As the temperature is annealed $\omega \rightarrow \infty$, the prior in Eq.~\eqref{eq:generative_graph_prior} collapses onto the DAG space and $G_\infty(\mathbf{Z}_k)$ defines an adjacency matrix. 

The prior for the latent graph embeddings $\mathbf{Z}_k$ is the product of four terms, each of which ensures specific properties of the generative graph prior. Namely, 
\begin{alignat}{2}\label{eq:latent_graph_prior}
    &p_{\beta, \omega}(\mathbf{Z}_k\mid \mathcal{D}_S)
    \\&\propto p\left(\mathcal{D}_{S,k} \mid G_\omega(\mathbf{Z}_k)\right) \notag\\
    &\quad\times \exp \left( -\beta\ \mathbb{E}_{p_\omega(G_k\mid \mathbf{Z}_k)}[h(G_k)]\right) \notag\\
    &\quad\times \prod_{i\not=j}\nolimits \mathcal{N}(\mathbf{U}^{(k)}_{\cdot i}\mid \bm{0}, \sigma_z I)\ \mathcal{N}(\mathbf{V}^{(k)}_{\cdot j}\mid \bm{0}, \sigma_z I) \notag\\
    &\quad\times p\left(G_\omega(\mathbf{Z}_k)\right)\notag.
\end{alignat}
The first term is our proposed \textit{elicited informative graph prior}, ensuring that the obtained embedding aligns with the expert's beliefs (see \Cref{sec:informative_elicitation_graph_prior}).
The last three terms, introduced by \citet{lorchDiBSDifferentiableBayesian2021a}, respectively constitute: a term that penalizes cyclicity in samples from the generative graph prior, priors ensuring that the inner products $\mathbf{U}^{(k)}_{\cdot i} \cdot \mathbf{V}^{(k)}_{\cdot j}$ remain ``well-behaved'', and a term that promotes specified random graph structures. Specifically, for Erd\H{o}s-Rényi graphs, \citet{lorchDiBSDifferentiableBayesian2021a} suggest the prior $p\left(G_\omega(\mathbf{Z}_k)\right) \propto q^{\lVert 
G_\omega(\mathbf{Z}_k) \rVert_1 } (1-q)^{\binom{d}{2} - \lVert G_\omega(\mathbf{Z}_k) \rVert_1}$, while scale-free graphs are promoted using $p\left(G_\omega(\mathbf{Z}_k)\right) \propto \prod_{i=1}^d \left(1 + \lVert  G_\omega(\mathbf{Z}_k)_i \rVert_1 \right)^{-3}$, where $G_\omega(\mathbf{Z}_k)_i$ is the $i^{\textrm{th}}$ row in the soft graph.  In Appendix~\ref{sec:vamsl-algorithm} we discuss implementation details with a full procedure to approximate the posterior distribution of $p(\pi,\mathbf{Z},G,\Theta,C\mid \mathcal{D})$. Algorithm~\ref{alg:edge-elicitation-vamsl} describes how to incorporate the expert's edge-beliefs to the approach described in this section.
\begin{algorithm}
\caption{VaMSL~+~expert}\label{alg:edge-elicitation-vamsl}
\begin{algorithmic}[1]
\Require Number of queries $M\times B$; observations ${\mathcal{D}=\{\mathbf{x}_n\}_{n=1}^N}$; number of components $K$
\State Obtain $q_{\beta,\omega}(\pi,\mathbf{Z},G,\Theta,C)$ that approximates $p(\pi,\mathbf{Z},G,\Theta,C\mid \mathcal{D})$, using Algorithm~\ref{alg:VaMSL}.
\State Choose the top $M$ queries for each of the $K$ components that maximizes the EIG (see: Section~\ref{sec:elicitation_BED}).\label{line:2}
\State Query the edge beliefs from the expert ($\mathcal{D}_S, \mathcal{D}_H$).
\State Update $q_{\beta,\omega}(\pi,\mathbf{Z},G,\Theta,C)$ to approximate $p(\pi,\mathbf{Z},G,\Theta,C\mid \mathcal{D},{\mathcal{D}_S})$, using Algorithm~\ref{alg:VaMSL}.
\State Go to Step~\ref{line:2} until out of query budget. 
\end{algorithmic}
\end{algorithm}

\section{NUMERICAL EXPERIMENTS AND CANCER LABELING} 
\label{sec:results}
We demonstrate that incorporating soft constraints through our proposed prior improves the inference, even when the expert can make mistakes. For this, we present experiments on single-component, two-component, and a breast-cancer data set. The experiments compare four approaches: VaMSL with expert information, VaMSL without expert information, a Gaussian mixture model \citep[GMM; using \texttt{scikit-learn},][]{pedregosaScikitlearnMachineLearning2011}, a causal k-means kernel \citep{markhamDistanceCovariancebasedKernel2022}, DiBS without label information, and DiBS with label information. We remark that the last approach is an ideal scenario (the labels ensure it has the correct clusters), as such it provides an unrealistic bound which no unsupervised method should be able to accomplish. 
In principle we could compare to the approach by \citet{marchantCovariateDependentMixture2025}, but we did not find an implementation of their method.
\begin{figure}[!htp]
  \centering
  \subfloat{\includegraphics[width=0.48\textwidth]{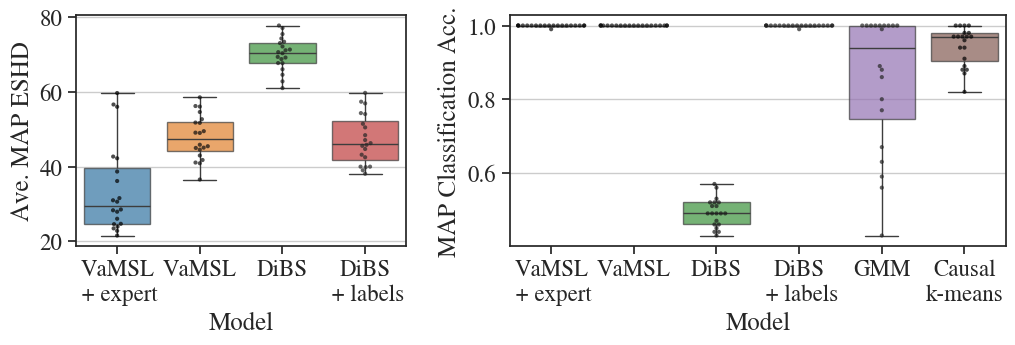}\label{fig:f2}}
  \hfill
  \subfloat{\includegraphics[width=0.48\textwidth]{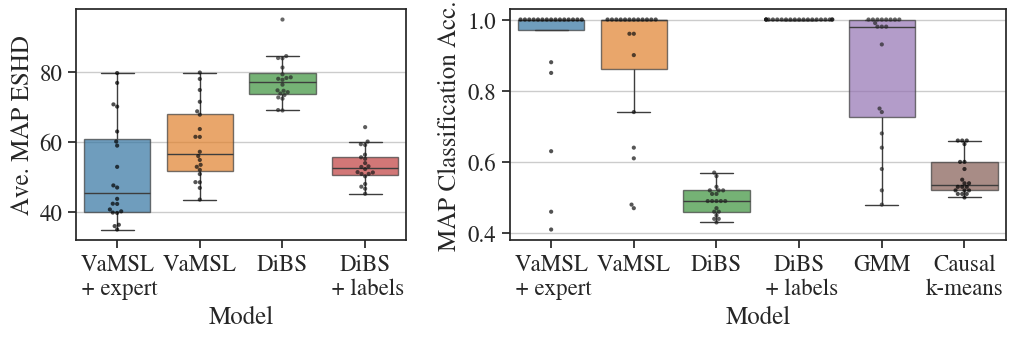}}
  \caption{Boxplot (and values) of average ESHD between learned graphs and true graphs (\textit{left}) and classification accuracy (\textit{right}) by method, in data generated from a mixture of two \textbf{linear} (\textit{top}) and \textbf{non-linear} (\textit{bottom}) ER BNs with Gaussian errors. Learned graphs not available for Causal k-means and GMMs.}
  \label{fig:synthetic_ER}
\end{figure}
\subsection{Data Generation and Metrics}
To validate VaMSL on synthetic settings, we generate data from Gaussian BNs with Erd\H{o}s-R\'enyi (ER) ground truth graphs parametrized by linear or non-linear mean functions. These graphs are generated with an expectation of two edges per node. Non-linearity is instantiated using shallow 2-layer neural networks (NN) with 5 neurons per hidden layer. All the parameters for the NNs and linear models are randomly set with a standard normal distribution.

We assess the performance of all methods in terms of classification accuracy (for correct labeling) with the \textit{maximum a posteriori} (MAP) component on held-out observations ($\mathcal{D}_{ho}$) as the percentage of correctly classified observations. When a graph is learned by the method (namely, DiBS and VaMSL variants), we assess the structure learning by the \textit{expected structural Hamming-distance} (ESHD), given by
\begin{equation*}
    \textrm{ESHD}(p(G_k\mid \mathcal{D}), G^*) =\ \mathbb{E}_{p(G_k\mid \mathcal{D})}\left[ \textrm{SHD}(G_k, G^*)\right],
\end{equation*}
where $\textrm{SHD}(G_k,G^*)$ is the number of edge changes needed to move from $G_k$ to $G^*$.
Additionally, in Appendix~\ref{app:numerical-experiments} we provide sensitivity studies, simulations with scale-free priors, other simulation results, and assessment of predictive power where the same conclusion of this section is echoed: learning the graphs by incorporating an expert helps the inference process compared to the competing baselines. An implementation of our algorithm is freely available in \href{https://github.com/ZacBjo/VaMSL_proj}{https://github.com/ZacBjo/VaMSL\_proj}.

Further ablation studies are provided in Appendices~\ref{app:expert-model}, \ref{app:scalability}, and \ref{app:component-misspecification}, where we compare against a hard-encoding strategy \citep[using the method of][]{sun_nts-notears_2023}, the scalability of our method to higher dimensions, and the behaviour with a misspecified number of components, respectively.
 
\subsection{Validation with Mixture of Graphs}\label{sec:hetergenous_experiments}
To investigate the 
heterogeneous setting we generate a two-component dataset $\mathcal{D} = \{\mathbf{x}_n \mid \mathbf{x}_n\in \mathbb{R}^{20}\}_{n=1}^{100}$ with mixing probabilities $(0.5,0.5)$, and each DAG is generated from the random graph prior. We repeat the simulations 20 times and evaluate on out-of-sample simulations. We apply the competing methods to the generated datasets. In Fig.~\ref{fig:synthetic_ER} we show the effectivity of both VaMSL as well as VaMSL paired with expert information (labeled VaMSL + expert) via our proposed prior for both linear and nonlinear models, respectively; in contrast to the competing methods.

The results in Fig.~\ref{fig:synthetic_ER} indicate how DiBS is \textit{unable} to accurately describe the underlying DGP---given by the graph-structures, as it tries to find a single graph to describe two separate graphs. VaMSL, on the other hand, is able to cluster the observations into homogeneous subpopulations (right panel) and infer a graph posterior comparable to those achieved when knowing the labels (left panel). Further, the VaMSL approaches outperform the GMM and the causal k-means methods in terms of classification accuracy. We recall that the k-means and GMM methods do not have a structure learning component so we cannot assess their ESHD.
\subsection{Querying Strategy Validation}
\label{sec:results_homogeneous}
\begin{figure}[!htp]
  \centering
  \subfloat{\includegraphics[width=0.5\textwidth]{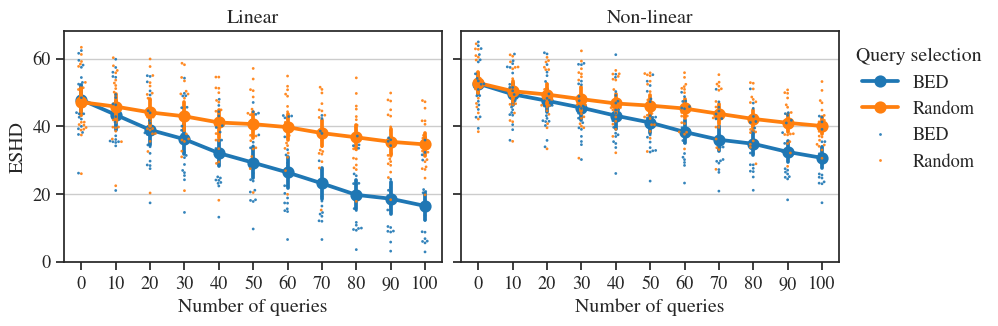}}
  \hfill
  \subfloat{\includegraphics[width=0.5\textwidth]{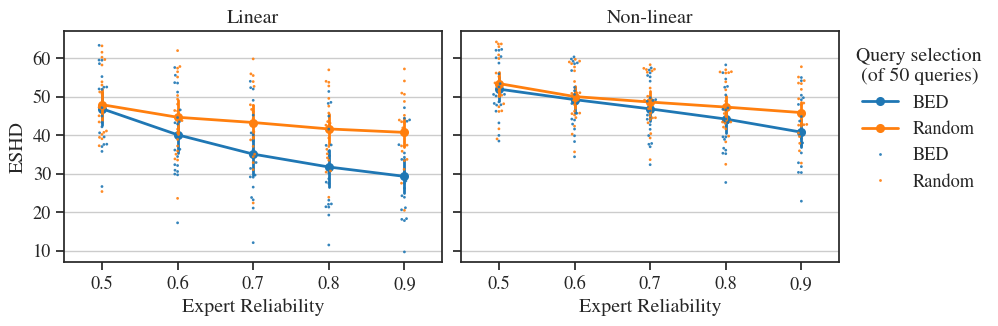}}
  \caption{Comparison between BED and random approaches for query selection by mean and bootstrapped 95\% confidence interval for homogeneous experiments with varying number of queries (\textit{top}) and varying expert reliability (\textit{bottom}) when inferring Gaussian ER network in the \textbf{linear} (\textit{left}) and \textbf{non-linear} case (\textit{right}).}
  \label{fig:querying_strategy_and_reliability}
\end{figure}

We consider a single-component setting, with different querying strategies and graph parametrizations (linear and nonlinear), for validation of our expert elicitation strategy (Section~\ref{sec:elicitation_BED}). 
Fig.~\ref{fig:querying_strategy_and_reliability}~(top) shows how increasing the number of queries to an expert (90\%~reliable) leads to improved structure learning, as expected. Additionally, our BED strategy outperforms random querying of an expert. 
In Fig.~\ref{fig:querying_strategy_and_reliability}~(bottom) we display the effect of the reliability of the expert, showing that more reliable opinions impact the inference more dramatically. As an expert would be the only one to reject an inferred model (regardless of predictive performance), it is indeed desirable that our prior emphasizes expert opinion proportionally to how strongly said opinion is held. In Appendix~\ref{app:expert-simulation} we detail how the expert responses are simulated.

\subsection{Label Assignment in Breast Cancer Data}\label{sec:application}
\begin{figure}
    \centering
    \includegraphics[width=\linewidth]{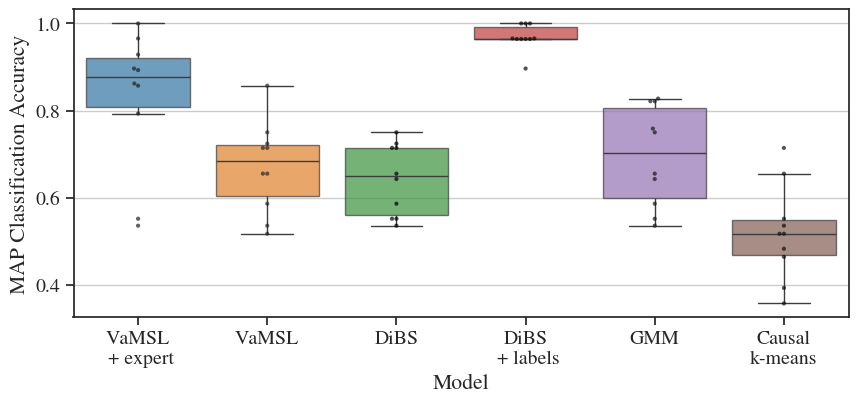}
    \caption{Boxplot and values of classification accuracy by method in out-of-sample data for the breast cancer dataset, in ten splits of the data.}
    \label{fig:UCI_clustering}
\end{figure}

As an application we use the Diagnostic Wisconsin Breast Cancer Database\footnote{from: \url{https://archive.ics.uci.edu/dataset/17/breast+cancer+wisconsin+diagnostic}.}. The data set consists of 569 observations with binary labels (for classification) and 30 real-valued features. We use the labels as the mixture components and interpret the problem as an unsupervised problem, with task to infer the labels.
As Fig.~\ref{fig:UCI_clustering} shows, VaMSL+expert outperforms all the other approaches, and VaMSL alone performs as good as a GMM. We provide details of the expert simulation as well as other relevant metrics in Appendix~\ref{app:cancer-details}. In Apppendix~\ref{app:cancer-expert-queries} we showcase how the performance of the method improves as the number of queries to the expert increases. We simulate the expert with a reliability of 0.9, more details in Appendix~\ref{app:expert-simulation}.

\section{CONCLUSIONS
}\label{sec:conclusions}
We have presented a strategy for eliciting expert beliefs about causal relationships as well as an elicited informative graph prior. Further, we have addressed the heterogeneous setting by proposing a mixture method, which when paired with the expert responses in the form of non-exchangeable component-wise priors, infers mixtures of CBNs. The method has been demonstrated to improve the structure learning and allows finding correct clusters with an unsupervised method, through an expert-in-the-loop.

We consider that our work can be expanded in at least four main directions. First, the proposed method we call ``VaMSL''  uses DiBS for the soft-graph inference part. This leaves open the possibility of using other graph discovery methods within VaMSL---different from DiBS, such as by \citet{annadaniBayesDAGGradientBasedPosterior2023}, \citet{charpentierDifferentiableDAGSampling2022}, and \citet{tranDifferentiableBayesianStructure2023a}, to encode the soft-graph distribution. Second, the assignment probabilities do not depend on covariates, a way to drop this assumption is to use a similar approach as \citet{marchantCovariateDependentMixture2025}. Third, \citet{darvariuLargeLanguageModels2024} propose querying a large language model for expert information, when an expert is not available, an idea which is easy to adapt to our proposed method. Finally, although our work is primarily concerned with the low data-regime (where prior information is most valuable), ensuring that our approach can scale to high dimensions --- where causal discovery algorithms become prohibitively expensive --- is of interest and is an active area of research in Bayesian causal discovery.


\subsubsection*{Acknowledgements}
This work was supported by the Research Council of Finland (Flagship programme: Finnish Center for Artificial Intelligence FCAI,~358958 and~359567), ELISE Networks of Excellence Centres (EU Horizon:~2020 grant agreement~951847) and UKRI Turing AI World-Leading Researcher Fellowship, EP/W002973/1. We acknowledge the computational resources provided by the Aalto Science-IT Project from Computer Science IT. We thank Yasir Barlas, Sammie Katt, and reviewers for comments on early versions of the manuscript. 




\bibliography{zotero_library} 

\section*{Checklist}
\begin{enumerate}

  \item For all models and algorithms presented, check if you include:
  \begin{enumerate}
    \item A clear description of the mathematical setting, assumptions, algorithm, and/or model. [Yes]
    \item An analysis of the properties and complexity (time, space, sample size) of any algorithm. [Yes, see Appendix~\ref{sec:vamsl-algorithm}.]
    \item (Optional) Anonymized source code, with specification of all dependencies, including external libraries. [Yes, our source code is available at \href{https://github.com/ZacBjo/VaMSL_proj}{https://github.com/ZacBjo/VaMSL\_proj}, and included in the Supplementary Material.]
  \end{enumerate}

  \item For any theoretical claim, check if you include:
  \begin{enumerate}
    \item Statements of the full set of assumptions of all theoretical results. [Yes]
    \item Complete proofs of all theoretical results. [Yes]
    \item Clear explanations of any assumptions. [Yes]     
  \end{enumerate}

  \item For all figures and tables that present empirical results, check if you include:
  \begin{enumerate}
    \item The code, data, and instructions needed to reproduce the main experimental results (either in the supplemental material or as a URL). [Yes]
    \item All the training details (e.g., data splits, hyperparameters, how they were chosen). [Yes]
    \item A clear definition of the specific measure or statistics and error bars (e.g., with respect to the random seed after running experiments multiple times). [Yes, we use boxplots for uncertainty quantification, instead of simple error bars.]
    \item A description of the computing infrastructure used. (e.g., type of GPUs, internal cluster, or cloud provider). [Yes, experiments were run on an internal cluster with NVIDIA H100 GPUs.]
  \end{enumerate}

  \item If you are using existing assets (e.g., code, data, models) or curating/releasing new assets, check if you include:
  \begin{enumerate}
    \item Citations of the creator If your work uses existing assets. [Yes]
    \item The license information of the assets, if applicable. [Not Applicable]
    \item New assets either in the supplemental material or as a URL, if applicable. [Yes]
    \item Information about consent from data providers/curators. [Not Applicable]
    \item Discussion of sensible content if applicable, e.g., personally identifiable information or offensive content. [Not Applicable]
  \end{enumerate}

  \item If you used crowdsourcing or conducted research with human subjects, check if you include:
  \begin{enumerate}
    \item The full text of instructions given to participants and screenshots. [Not Applicable]
    \item Descriptions of potential participant risks, with links to Institutional Review Board (IRB) approvals if applicable. [Not Applicable]
    \item The estimated hourly wage paid to participants and the total amount spent on participant compensation. [Not Applicable]
  \end{enumerate}

\end{enumerate}

\newpage
\onecolumn
\begin{appendices}
\aistatstitle{Incorporating Expert Knowledge into Bayesian Causal Discovery of Mixtures of Directed Acyclic Graphs: \\
Supplementary Materials}

\section{APPROXIMATING FAMILIES OF DISTRIBUTIONS}\label{sec:approximating-families}
We split the approximation of $p$ into three separate parts: approximate the component assignment ($\mathbf{c}_n$), the mixture probabilities ($\bm{\pi}$), and on each component the latent embeddings with the parameters ($\mathbf{Z}_k,\Theta_k$). First, to approximate the component assignment we use a categorical distribution, initialized as uninformative. Next, for the mixture probabilities we use a Dirichlet distribution with prior parameter $\alpha$, with a default value of $1$. Finally, to approximate the embeddings and parameters we use a particle approximation, as the one done by \citet{lorchDiBSDifferentiableBayesian2021a}.

\subsection{Description of VaMSL}\label{sec:vamsl-algorithm}
Algorithm~\ref{alg:VaMSL} describes the procedure we employ for learning mixtures of structures. The algorithm relies on six equations, which we provide next. The full derivation of these equations is given in Appendix~\ref{app:derivations-vamsl}. The algorithm consists of two different types of updates, which is enabled by a coordinate-ascent procedure. First, the updates on the mixing probabilities ($\bm{\pi}$) and assignment probabilities ($q(\mathbf{c}_n)$)---also termed responsibilities. These updates can be done explicitly as we describe in Appendix~\ref{app:mixing-updates}. Next, the updates of the latent embeddings ($\mathbf{Z}_k$) --- which encode soft graphs and the parameters ($\Theta_k$). These updates are done using SVGD, and fully described in Appendix~\ref{app:svgd-updates}. 

Denoting by $\zeta$ the digamma function, the update of the responsibilities $q(\mathbf{c}_n)$ is given by
\begin{align}
    \hat{q}^*(c_{nk}=1) = &\frac{\exp\left(\frac{1}{P}\sum_p\left[\ln p(\mathbf{x}_n\mid  G_k^{(p)}, \Theta_k^{(p)})\right] +  \zeta(\alpha_k) - \zeta(\sum_h^K\nolimits \alpha_h)\right)}{\sum_{h=1}^K \exp\left(\frac{1}{P}\sum_p\left[\ln p(\mathbf{x}_n\mid G_h^{(p)}, \Theta_k^{(p)})\right] + \zeta(\alpha_h) - \zeta(\sum_m^K\nolimits \alpha_m)\right)}. \label{eq:vamsl_update_responsibilities}
\end{align}

The mixing weights are updated according to
\begin{align}
        \ln q^*(\bm{\pi}) = \sum_k\nolimits \ln\pi_k\left[\left(\alpha_k + \sum_n\nolimits\hat{q}^*(c_{nk}=1)\right) - 1\right] + \textrm{constant}\label{eq:vamsl_update_mixing}.
\end{align}

The particle updates within each component $k=1,\dots,K$, are
\begin{align}
   \mathbf{Z}_k^{(p), t+1} &=\mathbf{Z}_k^{(p), t} + \eta_t \phi^Z_t(\Gamma_k^{(p),t}) \label{eq:vamsl_update_z_particle}\\
    \Theta_k^{(p), t+1} &= \Theta_k^{(p), t} + \eta_t \phi^\Theta_t(\Gamma_k^{(p),t}), \label{eq:vamsl_update_theta_particle} 
\end{align}
where
\begin{align}
    \phi^Z_t(\Gamma) &:= \frac{1}{P} \sum_p \left[\kappa(\Gamma_k^{(p),t}, \Gamma)  \nabla_{Z_k^{(p), t}}\ln q(\mathbf{Z}_k^{(p), t}, \Theta_k^{(p), t}) + \nabla_{Z_k^{(p), t}}\kappa(\Gamma_k^{(p),t}, \Gamma) \right], \label{eq:vamsl_z_particle_stein_functions}\\
    \phi^\Theta_t(\Gamma) &:= \frac{1}{P} \sum_p \left[\kappa(\Gamma_k^{(p),t}, \Gamma) \nabla_{\Theta_k^{(p), t}}\ln q(\mathbf{Z}_k^{(p), t}, \Theta_k^{(p), t}) + \nabla_{\Theta_k^{(p), t}}\kappa(\Gamma_k^{(p),t}, \Gamma)\right], \label{eq:vamsl_theta_particle_stein_functions}
\end{align}
using $\kappa$ as a squared exponential additive kernel and $\Gamma_k^{(p),t} = (\mathbf{Z}_k^{(p), t},\Theta_k^{(p),t})$.

\begin{algorithm}[!htb]
\caption{Variational Mixture Structure Learning} \label{alg:VaMSL}
\begin{algorithmic}[1]
\Require Observations $\mathcal{D} = \{\mathbf{x}_n\}_{n=1}^N$, with $\mathbf{x}_n\in\mathbb{R}^d$, number of components $K$, expert information~$\mathcal{D}_{\Psi}$ 

\hspace{-0.9cm} \textbf{Hyperparameters:} mixing prior $\bm{\alpha}$, kernel $\kappa$, schedules for $\omega_t, \beta_t$, and $\eta_t$, number of optimization steps $T$, number of coordinate-ascent updates $U$. 

\hspace{-0.9cm} \textbf{Output:} Approximate distribution of $\left\{\mathbf{Z}_k,\Theta_{k}\right\}_{k=1}^{K}$; distributions $\textrm{q}(C)$ and $\textrm{q}(\bm{\pi})$.
\State Initialize particles $\{\mathbf{Z}_k^{(p),0}, \Theta_k^{(p),0}\}_{p=1}^P$
\For{$u=1$ until $U+1$}\Comment{Coordinate ascent}
    \State Update responsibilities $\textrm{q}(\bm{c}_n)$ using Equation~\eqref{eq:vamsl_update_responsibilities}, for $n=1,\dots,N$. \label{line:3} 
    \State Update mixing weights $\textrm{q}(\bm{\pi})$ using Equation~\eqref{eq:vamsl_update_mixing}. 
    \State Sample assignments $\bm{c}_n' \sim \textrm{q}(\bm{c}_n),\ n=1,\ldots,N$.
    \For{$k=1$ until $K$}
        \For{$t=1$ until $T+1$} \Comment{Optimization}
                \State Obtain particles $\{\mathbf{Z}_k^{(p),t+1},\Theta_{k}^{(p),t+1}\}_{p=1}^P$ using Equations~\eqref{eq:vamsl_update_z_particle}-\eqref{eq:vamsl_theta_particle_stein_functions}.
        \EndFor 
    \EndFor
\EndFor \label{line:11}
\end{algorithmic}
\end{algorithm}

Additionally, since VaMSL is dependent on initialization and component collapsing is possible, we include random restarts to ensure better results. We set two conditions for randomly restarting with reinitialized particles, 1) if updating the responsibilities (line~\ref{line:3}) numerically collapses one of the components to have zero responsibility, and 2) if after the final annealing of the particle distribution (line~\ref{line:11}) none of the graphs $G_\infty(\mathbf{Z}_k^{(p)}), p=1,\dots, P,$ for a given component $k$ are acyclic. We set the maximum number of random restarts to $5$ restarts, and, in the case of using an elicited informative prior, let the reinitialized model start with same prior.

The dominating step of our SVGD instantiation of VaMSL lies in computing the (approximate) gradients part of the SVGD updates in \Cref{eq:vamsl_z_particle_stein_functions} and \Cref{eq:vamsl_theta_particle_stein_functions}. This computational load can be addressed by instead computing the gradients using mini-batches, thereby avoiding a dominating complexity term dependent on the number of observations $N$. The SVGD updates also involve evaluating kernels which depend on the number of mixture components $K$, the number of particles $P$, number of covariates $d$, and the latent dimensionality $l$, and have the complexity $\mathcal{O}(KP^2dl)$. We also note that when the update of the responsibilities (given in \Cref{eq:vamsl_update_responsibilities}) is carried out for each observation (as in our experiments) it scales linearly in $N$ and depends on the number or components $K$ and particles $P$, with a complexity of  $\mathcal{O}(NKP)$.

\subsection{Proof of Equalities Needed for VaMSL}\label{app:derivations-vamsl}
As indicated in Appendix~\ref{sec:vamsl-algorithm}, the Equations~\eqref{eq:vamsl_update_responsibilities}-\eqref{eq:vamsl_theta_particle_stein_functions} are sufficient because of CAVI updates. For in depth explanations of CAVI, see \citet{bleiVariationalInferenceReview2017} or \citet[Chapter 10]{bishopPatternRecognitionMachine2006}. The optimal settings (denoted by $q^*$) for the individual variational distributions, conditional on values of the other components are: 
\begin{align}
    \ln q^*(\mathbf{Z}_k, \Theta_k) &= \underset{q\in\mathcal{Q}}{\mathrm{argmax}} \hspace{1mm} \mathbb{E}_{q(\bm{\pi})q(C)q(Z_{\neg k}, \Theta_{\neg k})}[\ln p(\mathbf{Z}, \Theta, C, \bm{\pi}, \mathcal{D} \mid \mathcal{D}_S, \mathcal{D}_H)], \label{eq:CAVI_Z_Theta} \\
    \ln q^*(\bm{c}_{n}) &= \underset{q\in\mathcal{Q}}{\mathrm{argmax}} \hspace{1mm} \mathbb{E}_{q(\bm{\pi})q(C_{\neg n})q(\mathbf{Z}, \Theta)}[\ln p(\mathbf{Z}, \Theta, C, \bm{\pi}, \mathcal{D} \mid \mathcal{D}_S, \mathcal{D}_H)]\text{, and} \label{eq:CAVI_C} \\
    \ln q^*(\bm{\pi}) &= \underset{q\in\mathcal{Q}}{\mathrm{argmax}} \hspace{1mm} \mathbb{E}_{q(C)q (\mathbf{Z}, \Theta)}[\ln p(\mathbf{Z}, \Theta, C, \bm{\pi}, \mathcal{D} \mid \mathcal{D}_S, \mathcal{D}_H)],\label{eq:CAVI_pi}
\end{align}
where $k=1,\ldots,K$, $n=1,\ldots,N$, and the indexing $\neg i$ refers to a product between all the distributions with indices $j \neq i$. As these terms are dependent on each other, we employ CAVI and Monte Carlo estimates to iteratively refine our variational approximations. Next we describe the computation of these terms, see \Cref{sec:CAVI_derivations} for full derivations of each CAVI update.

For our approach to be valid, we make use of Proposition 1 and Equation~(16) of \citet{lorchDiBSDifferentiableBayesian2021a} to ensure convergence of the samples we obtain with the CAVI updates. For completeness, we include them below, adapted to our notation.
\begin{proposition}[Latent posterior expectation,  \citet{lorchDiBSDifferentiableBayesian2021a}]
Under the generative model~\eqref{eq:generative_graph_prior}, with $K=1$, it holds that:
\begin{enumerate}
    \item $\mathbb{E}_{p(G\mid \mathcal{D})}[f(G)] = \mathbb{E}_{p(\mathbf{Z}\mid \mathcal{D})}\left[\frac{\mathbb{E}_{p(G\mid \mathbf{Z})}[f(G)p(\mathcal{D}\mid G)] }{\mathbb{E}_{p(G\mid \mathbf{Z})}[p(\mathcal{D}\mid G)] }\right]$ and
    \item $\mathbb{E}_{p(G,\Theta \mid \mathcal{D})}[f(G,\Theta)] = \mathbb{E}_{p(\mathbf{Z},\Theta\mid \mathcal{D})}\left[\frac{\mathbb{E}_{p(G\mid \mathbf{Z})}[f(G,\Theta)p(\Theta\mid G)p(\mathcal{D}\mid G,\Theta)] }{\mathbb{E}_{p(G\mid \mathbf{Z})}[p(\Theta\mid G)p(\mathcal{D}\mid G,\Theta)] }\right]$.
\end{enumerate}
\end{proposition}

Additionally, Equation~(16) of \citet{lorchDiBSDifferentiableBayesian2021a} indicates that when annealing $\omega\to\infty$ we obtain:
\begin{align}
    \mathbb{E}_{p(G\mid \mathcal{D})}[f(G)] &\to \mathbb{E}_{p(\mathbf{Z}\mid \mathcal{D})}[f(G_\infty(\mathbf{Z}))] \nonumber \\
    \mathbb{E}_{p(G,\Theta\mid \mathcal{D})}[f(G,\Theta)] &\to \mathbb{E}_{p(\mathbf{Z},\Theta\mid \mathcal{D})}[f(G_\infty(\mathbf{Z}),\Theta)] \label{eq:Lorch_16}.
\end{align}
As a note, the equation numbers came out in a nice enough way that our label of Eq.~\eqref{eq:Lorch_16} matches the label of Eq.~(16) in~\citet{lorchDiBSDifferentiableBayesian2021a}. We make use of this coincidence.

Combining these results allows working with the annealed version of the soft-graph to obtain estimates on functions that depend on the graph $G$. However, we want to be able compute expectations with respect to latent graphs also in the mixture setting. We therefore also prove an equality for computing expectations in the graph space via latent graph embeddings in the mixture setting in Proposition 2.

\begin{proposition}[Latent mixture posterior expectation]\label{prop:LMPE}
Under the generative model~\eqref{eq:generative_graph_prior}, it holds that:
\begin{equation*}
    \mathbb{E}_{p(G_k,\Theta_k \mid \mathcal{D})}[f(G,\Theta)] = \mathbb{E}_{p(\mathbf{Z}_k,\Theta_k, C \mid \mathcal{D})}\left[\frac{\mathbb{E}_{p(G_k \mid \mathbf{Z}_k)}[f(G_k,\Theta_k)p(\Theta_k \mid G_k)\prod_n p(\mathbf{x}_n\mid G_k,\Theta_k)^{c_{nk}}] }{\mathbb{E}_{p(G_k\mid \mathbf{Z}_k)}[p(\Theta_k \mid G_k)\prod_n p(\mathbf{x}_n\mid G_k,\Theta_k)^{c_{nk}}] }\right].
\end{equation*}
\end{proposition}

We provide a proof in Appendix~\ref{app:latent_mixture_posterior_expectation_derivation}.

\subsubsection{Exact Update of Responsibilities and Mixing Weights}\label{app:mixing-updates}
Making use of Equation~\eqref{eq:Lorch_16} \citep{lorchDiBSDifferentiableBayesian2021a}, the optimized distributions $q(\mathbf{Z}, \Theta)$ can be rewritten as distributions $q(G, \Theta)$ over CBNs. The optimal update given in \Cref{eq:CAVI_C} is simplified to 
\begin{equation*}
    \ln q^*(\bm{c}_n) = \mathbb{E}_{q(G,\Theta)}\left[\ln p(\mathbf{x}_n \mid G, \Theta, \bm{c}_n))\right] + \mathbb{E}_{q(\bm{\pi})}\left[\ln p(\mathbf{c}_n \mid \bm{\pi})\right] + \textrm{constant},
\end{equation*}
and the optimal variational distribution for the assignment variables given current approximations $q(G, \Theta)q(\bm{\pi})$ satisfies
\begin{equation}
    q^*(\bm{c}_n) \propto \prod_k\nolimits \exp\left(\mathbb{E}_{q(G_k,\Theta_k)}\left[\ln p(\mathbf{x}_n\mid G_k, \Theta_k)\right] +  \mathbb{E}_{q(\bm{\pi})}\left[\ln\pi_k\right]\right)^{c_{nk}}, n=1,\ldots,N.
\end{equation}
Normalizing across the components implies 
\begin{equation}\label{eq:variational_C_update_marginal}
    q^*(c_{nk} = 1) = \frac{\exp\left(\mathbb{E}_{q(G_k,\Theta_k)}\left[\ln p(\mathbf{x}_n \mid G_k, \Theta_k)\right] +  \mathbb{E}_{q(\bm{\pi})}\left[\ln\pi_k\right]\right)}{\sum_{h=1}^K \exp\left(\mathbb{E}_{q(G_h,\Theta_h)}\left[\ln p(\mathbf{x}_n\mid  G_h, \Theta_h)\right] +  \mathbb{E}_{q(\bm{\pi})}\left[\ln\pi_h\right]\right)}.
\end{equation}
Since $q(\bm{\pi})$ is a density of Dirichlet random variable (see Appendix~\ref{sec:approximating-families}), use the standard result for the expectation of the logarithm a Dirichlet distributed variable: $\mathbb{E}_{q(\bm{\pi})}\left[\ln\pi_h\right] = \zeta(\alpha_k) - \zeta(\sum_h^K\nolimits \alpha_h)$, where $\zeta(x) = \frac{\partial}{\partial x} \Gamma(x)$ is the digamma function. The expectation $\mathbb{E}_{q(G_k,\Theta_k)}$, appearing in both the numerator and denominator of \Cref{eq:variational_C_update_marginal}, is approximated with a Monte Carlo average
\begin{align}
    \hat{q}^*(c_{nk}=1) = &\frac{\exp\left(\frac{1}{P}\sum_p\left[\ln p(\mathbf{x}_n\mid G_k^{(p)}, \Theta_k^{(p)})\right] +  \zeta(\alpha_k) - \zeta(\sum_h^K\nolimits \alpha_h)\right)}{\sum_{h=1}^K \exp\left(\frac{1}{P}\sum_p\left[\ln p(\mathbf{x}_n\mid G_h^{(p)}, \Theta_h^{(p)})\right] + \zeta(\alpha_h) - \zeta(\sum_m^K\nolimits \alpha_m)\right)}. \label{eq:vamsl_approximate_assignment_update}
\end{align}
We also note that, the expectation in \Cref{eq:variational_C_update_marginal} with respect to the distribution $q(G_k,\Theta_k)$ requires the latent graph embeddings to be annealed as described above when presenting Proposition 1. If however, the assignments are updated before sufficient annealing, Proposition~\ref{prop:LMPE}  can be employed to compute the expectations via the distributions $q(Z_k,\Theta_k)$. For our algorithms, we use Proposition~\ref{prop:LMPE} in this manner when updating the responsibilities, except after the final annealing of the particle distribution $q(Z_k,\Theta_k)$, in which case we compute \Cref{eq:variational_C_update_marginal} using hard graphs from $q(G_{\infty}(Z_k),\Theta_k)$.


Due to our choice of prior $p(\bm{\pi})=\textrm{Dir}(\bm{\alpha})$, the update of the variational distribution $q(\bm{\pi})$, given current distributions $q(\mathbf{Z}, \Theta)q(C)$, is a standard result \citep[see, for instance,][Chapter 10]{bishopPatternRecognitionMachine2006} given by:
\begin{equation} \label{eq:vamsl_mixing_weights_update}
    \ln q^*(\bm{\pi}) = \sum_k\nolimits \ln\pi_k\left[\left(\alpha_k + \sum_n\nolimits\hat{q}^{*}(c_{nk}=1)\right) - 1\right] + \textrm{constant}.
\end{equation}
The optimal update is, therefore, also Dirichlet distributed and $q^*(\bm{\pi}) = \textrm{Dir}(\bm{\alpha}^*)$, where $ \alpha_k^* = \alpha_k + \sum_n\nolimits \hat{q}^{*}(c_{nk}=1)$.

\subsubsection{Particle Algorithm for Latent Embeddings and Parameters}\label{app:svgd-updates}
To approximate full posteriors using SVGD, we instantiate a set of $P$ particles $\{\mathbf{Z}_k^{(p)}, \Theta_k^{(p)}\}_{p=1}^P$ for each component, $k=1,\ldots,K$, and perturb each in the direction maximally decreasing the Kullback-Leibler divergence. Using $\Gamma_k^{(p),t} = (\mathbf{Z}_k^{(p), t}, \Theta_k^{(p), t})$, for steps $t=1,\ldots,T$, we apply the following perturbation to each particle:
\begin{align}
   \mathbf{Z}_k^{(p), t+1} &=\mathbf{Z}_k^{(p), t} + \eta_t \phi^Z_t(\Gamma_k^{(p),t}) \label{eq:vamsl_z_particle_update}\quad \textrm{and}\quad \Theta_k^{(p), t+1} = \Theta_k^{(p), t} + \eta_t \phi^\Theta_t(\Gamma_k^{(p),t}), 
\end{align}
where $p=1,\ldots,P$, $k=1,\ldots,K$, and $\eta_t$ is a stepsize. The transforms $\phi_t^Z$ and $\phi_t^\Theta$ take the form:
\begin{align}
    \phi^Z_t(\cdot) &:= \frac{1}{P} \sum_p \left[\kappa(\Gamma_k^{(p),t}, \cdot)  \nabla_{Z_k^{(p), t}}\lnq(Z_k^{(p), t}, \Theta_k^{(p), t}) + \nabla_{Z_k^{(p), t}}\kappa(\Gamma_k^{(p),t}, \cdot) \right], \label{eq:vamsl_z_particle_stein_function}\\
    \phi^\Theta_t(\cdot) &:= \frac{1}{P} \sum_p \left[\kappa(\Gamma_k^{(p),t}, \cdot) \nabla_{\Theta_k^{(p), t}}\lnq(Z_k^{(p), t}, \Theta_k^{(p), t}) + \nabla_{\Theta_k^{(p), t}}\kappa(\Gamma_k^{(p),t}, \cdot)\right], \label{eq:vamsl_theta_particle_stein_function}
\end{align}
where the gradients are estimated using samples $\bm{c}^{(s)}_n \sim q(\bm{c}_n ), s=1,\ldots,S,n=1,\ldots,N$ as shown in \Cref{eq:vamsl_z_approx_gradient} and \Cref{eq:vamsl_theta_approx_gradient}. The kernel $\kappa$ needs to be positive-definite and \citet{lorchDiBSDifferentiableBayesian2021a} suggest using the additive \textit{SE} kernel:
\begin{equation}
    \kappa\left((\mathbf{Z}, \Theta), (\mathbf{Z}', \Theta')\right) := \exp\left(-\frac{\Vert\mathbf{Z}-\mathbf{Z}' \Vert_2^2}{\gamma_Z} \right) + \exp\left(-\frac{ \Vert \Theta-\Theta' \Vert_2^2}{\gamma_\Theta} \right),
    \label{eq:additive_SE_kernel}
\end{equation}
where $\gamma_Z$ and $\gamma_\Theta$ are length-scales.

Note that the particles $\mathbf{Z}_k^{(p)}$ can be interpreted as continuous relaxations for unique graphs $G_k^{(p)}$ as the temperature $\omega$ in \Cref{eq:generative_graph_prior} goes to infinity. To ensure that these graphs are DAGs, the temperature $\beta$ of the latent prior in \Cref{eq:latent_graph_prior} is annealed $\beta \rightarrow \infty$ over the course of the optimization as well. The updated distributions $q(\mathbf{Z}_k, \Theta_k), k=1,\ldots,K$ can thus be rewritten as distributions over CBNs: $q(G_k, \Theta_k), k=1,\ldots,K$, where each graph particle $G_k^{(p)}$ is a DAG and satisfies $p_{\infty}(G_k^{(p)}\mid \mathbf{Z}_k^{(p)}, E_k) = 1$, by alluding to Proposition~1 \citep{lorchDiBSDifferentiableBayesian2021a}.

\subsubsection{Derivations of CAVI Updates}
\label{sec:CAVI_derivations}
The results presented above depend on the gradients $\nabla_{Z_k}\ln q(\mathbf{Z}_k, \Theta_k)$, $\nabla_{Z_k^{(p)}}\ln q(Z_k^{(p)}, \Theta_k^{(p)})$, $\nabla_{\Theta_k}\ln q(\mathbf{Z}_k, \Theta_k)$, $\nabla_{\Theta_k^{(p)}}\ln q(Z_k^{(p)}, \Theta_k^{(p)})$, as well as the updates for the variational responsibilities $q(\bm{c}_n)$ and mixing weights $q(\bm{\pi})$. We present the derivations of these gradients below. Note that the CAVI updates for the responsibilities and mixing weights are calculated with respect to graph distributions $q(G_k, \Theta_k)$ as (continuous) latent embedding space maps unto the discrete graph space as the temperature in equation \eqref{eq:generative_graph_prior} is annealed, $\omega \to \infty$. We suppress the conditioning on expert information $\mathcal{D}_{S,k}, \mathcal{D}_{H,k}, k=1,\ldots,K$ for clarity.

\textbf{Component Latent Graph Embedding Gradient}
The gradient of $\ln q(\mathbf{Z}_k, \Theta_k)$ with respect to the latent graph embeddings is obtained by:
\begingroup
\allowdisplaybreaks 
\addtolength{\jot}{0.3em} 
\begin{align*}
    &\nabla_{Z_k}\ln q(\mathbf{Z}_k, \Theta_k) \\
    &= \nabla_{Z_k}\mathbb{E}_{ q(\bm{\pi}) q(C)\prod_{h \neq k} q(Z_h,  \Theta_h)}\left[\ln p(\mathbf{Z},\Theta, C, \bm{\pi}, \mathcal{D})\right]
    \\
    &=\begin{aligned}[t]
        &\nabla_{Z_k}\mathbb{E}_{ q(\bm{\pi}) q(C)\prod_{h \neq k} q(Z_h,\Theta_h)}\left[\ln p(\Theta, \mathcal{D}\mid\mathbf{Z}, C))\right] 
        && \text{(Expand generative model.)}
        \\
        &+ \nabla_{Z_k}\mathbb{E}_{ q(\bm{\pi}) q(C)\prod_{h \neq k} q(Z_h,\Theta_h)}\left[\ln p(\mathbf{Z})\right]
        \\
        &+ \nabla_{Z_k}\mathbb{E}_{ q(\bm{\pi}) q(C)\prod_{h \neq k} q(Z_h,\Theta_h)}\left[\ln p(C\mid \bm{\pi})\right]
        \\
        &+ \nabla_{Z_k}\mathbb{E}_{ q(\bm{\pi}) q(C)\prod_{h \neq k} q(Z_h,\Theta_h)}\left[\ln p(\bm{\pi})\right] 
    \end{aligned}
    \\
    &=\begin{aligned}[t]
        &\nabla_{Z_k}\mathbb{E}_{ q(\bm{\pi}) q(C)\prod_{h \neq k} q(Z_h,\Theta_h)}\left[\ln p(\Theta, \mathcal{D}\mid\mathbf{Z}, C)\right] 
        && \text{(Remove terms independent of $\nabla_{Z_k}$.)}
        \\
        &+ \nabla_{Z_k}\mathbb{E}_{ q(\bm{\pi}) q(C)\prod_{h \neq k} q(Z_h,\Theta_h)}\left[\ln p(\mathbf{Z})\right]
    \end{aligned}
    \\
    &=\begin{aligned}[t]
        &\nabla_{Z_k}\mathbb{E}_{ q(\bm{\pi}) q(C)\prod_{h \neq k} q(Z_h,\Theta_h)}\left[\sum_m\nolimits \ln p(\Theta_m, \mathcal{D}\mid Z_m, C)\right]
        \\
        &+ \nabla_{Z_k}\mathbb{E}_{ q(\bm{\pi}) q(C)\prod_{h \neq k} q(Z_h,\Theta_h)}\left[\sum_m\nolimits\ln p(Z_m)\right]
    \end{aligned} 
    \\
    &=\begin{aligned}[t]
        &\nabla_{Z_k}\mathbb{E}_{ q(\bm{\pi}) q(C)\prod_{h \neq k} q(Z_h,\Theta_h)}\left[\ln p(\Theta_k, \mathcal{D}\mid\mathbf{Z}_k, C)\right]
        && \text{(Terms where $m \neq k$ don't affect the gradient.)}
        \\
        &+ \nabla_{Z_k}\mathbb{E}_{ q(\bm{\pi}) q(C)\prod_{h \neq k} q(Z_h,\Theta_h)}\left[\ln p(\mathbf{Z}_k)\right]
    \end{aligned} 
    \\
    &=\begin{aligned}[t]
        &\nabla_{Z_k}\mathbb{E}_{\prod_{n} q(\bm{c}_n)}\left[\ln p(\Theta_k, \mathcal{D}\mid\mathbf{Z}_k, C)\right] + \nabla_{Z_k}\ln p(\mathbf{Z}_k)
        && \text{(Factor out independent expectations.)}
    \end{aligned}
    \\
    &=\begin{aligned}[t]
        &\nabla_{Z_k}\mathbb{E}_{\prod_{n} q(\bm{c}_n)}\left[ \ln\left[\sum_{G_k}\nolimits  p(G_k\mid\mathbf{Z}_k) p(\Theta_k, \mathcal{D}\mid G_k, C)\right]\right] 
        && \text{(From generative model.)}
        \\
        &+ \nabla_{Z_k}\ln p(\mathbf{Z}_k)
    \end{aligned}
    \\
    &=\begin{aligned}[t]
        &\mathbb{E}_{\prod_{n} q(\bm{c}_n)}\left[\nabla_{Z_k} \ln\left[ \mathbb{E}_{ p(G_k\mid \mathbf{Z}_k)}\left[ p(\Theta_k, \mathcal{D}
        \mid G_k, C)\right]\right]\right]
        \\
        &+ \nabla_{Z_k}\ln p(\mathbf{Z}_k)
    \end{aligned}
    \\
    &=\begin{aligned}[t]
        &\mathbb{E}_{\prod_{n} q(\bm{c}_n)}\left[\nabla_{Z_k} \ln\left[ \mathbb{E}_{ p(G_k\mid\mathbf{Z}_k)}\left[ p(\Theta_k\mid G_k) \prod_n\nolimits p(\mathbf{x}_n\mid G_k, \Theta_k)^{c_{nk}}\right]\right]\right]
        \\
        &+ \nabla_{Z_k}\ln p(\mathbf{Z}_k)
    \end{aligned}
    \\
    &=\begin{aligned}[t]
        &\mathbb{E}_{\prod_{n} q(\bm{c}_n)}\left[\frac{\nabla_{Z_k} \mathbb{E}_{ p(G_k\mid\mathbf{Z}_k)}\left[p(\Theta_k\mid G_k) \prod_n\nolimits p(\mathbf{x}_n\mid G_k, \Theta_k)^{c_{nk}}\right]}{ \mathbb{E}_{ p(G_k\mid \mathbf{Z}_k)}\left[p(\Theta_k\mid G_k) \prod_n\nolimits p(\mathbf{x}_n\mid G_k, \Theta_k)^{c_{nk}}\right]}\right]
        && \text{($\nabla \ln f(x) = \frac{\nabla f(x)}{f(x)}$.)}
        \\
        &+ \nabla_{Z_k}\ln p(\mathbf{Z}_k)
    \end{aligned}
\end{align*}
\endgroup
With samples $\{\bm{c}_n'\}_{n=1}^N, s=1,\ldots,S$ drawn from the variational distribution $ q(C)$, we can compute a Monte Carlo estimate of the gradient:
\begin{equation}
    \nabla_{Z_k}\ln q(\mathbf{Z}_k, \Theta_k) \approx \nabla_{Z_k}\ln p(\mathbf{Z}_k) + \frac{1}{S}\sum_s \frac{\nabla_{Z_k}\mathbb{E}_{p(G_k \mid \mathbf{Z}_k)}\left[p(\Theta_k\mid G_k) \prod_n\nolimits p(\mathbf{x}_n\mid  G_k, \Theta_k)^{c_{nk}^{(s)}}\right]}{\mathbb{E}_{p(G_k\mid\mathbf{Z}_k)}\left[p(\Theta_k\mid G_k) \prod_n\nolimits p(\mathbf{x}_n\mid G_k, \Theta_k)^{c_{nk}^{(s)}}\right]}. \label{eq:vamsl_z_approx_gradient}
\end{equation}

\textbf{Component Parameter Gradient}
The gradient of $\ln q(\mathbf{Z}_k, \Theta_k)$ with respect to the parameters is obtained by:
\begingroup
\allowdisplaybreaks 
\addtolength{\jot}{0.3em} 
\begin{align*}
    &\nabla_{\Theta_k}\ln q(\mathbf{Z}_k, \Theta_k)\\
    &= \nabla_{\Theta_k}\mathbb{E}_{ q(\bm{\pi}) q(C)\prod_{h \neq k} q(Z_h,\Theta_h)}\left[\ln p(\mathbf{Z},\Theta, C, \bm{\pi}, \mathcal{D})\right]
    \\
    &=\begin{aligned}[t]
        &\nabla_{\Theta_k}\mathbb{E}_{ q(\bm{\pi}) q(C)\prod_{h \neq k} q(Z_h,\Theta_h)}\left[\ln p(\Theta, \mathcal{D}\mid\mathbf{Z}, C))\right]
        && \text{({Expand generative model.})}
        \\
        &+ \nabla_{\Theta_k}\mathbb{E}_{ q(\bm{\pi}) q(C)\prod_{h \neq k} q(Z_h,\Theta_h)}\left[\ln p(\mathbf{Z})\right]
        \\
        &+ \nabla_{\Theta_k}\mathbb{E}_{ q(\bm{\pi}) q(C)\prod_{h \neq k} q(Z_h,\Theta_h)}\left[\ln p(C\mid \bm{\pi})\right]
        \\
        &+ \nabla_{\Theta_k}\mathbb{E}_{ q(\bm{\pi}) q(C)\prod_{h \neq k} q(Z_h,\Theta_h)}\left[\ln p(\bm{\pi})\right] 
    \end{aligned}
    \\
    &=\begin{aligned}[t]
        &\nabla_{\Theta_k}\mathbb{E}_{ q(\bm{\pi}) q(C)\prod_{h \neq k} q(Z_h,\Theta_h)}\left[\ln p(\Theta, \mathcal{D}\mid\mathbf{Z}, C)\right] 
        && \text{(Remove terms independent of $\nabla_{\Theta_k}$.)} 
    \end{aligned}
    \\
    &=\begin{aligned}[t]
        &\nabla_{\Theta_k}\mathbb{E}_{ q(\bm{\pi}) q(C)\prod_{h \neq k} q(Z_h,\Theta_h)}\left[\sum_m\nolimits \ln p(\Theta_m, \mathcal{D}\mid \mathbf{Z}_m, C)\right]
    \end{aligned}
    \\
    &=\begin{aligned}[t]
        &\nabla_{\Theta_k}\mathbb{E}_{ q(\bm{\pi}) q(C)\prod_{h \neq k} q(Z_h,\Theta_h)}\left[\ln p(\Theta_k, \mathcal{D}\mid\mathbf{Z}_k, C)\right]
        && \text{(Terms where $m \neq k$ do not affect the gradient.)}
    \end{aligned} 
    \\
    &=\begin{aligned}[t]
        &\nabla_{\Theta_k}\mathbb{E}_{\prod_{n} q(\bm{c}_n)}\left[\ln p(\Theta_k, \mathcal{D}\mid \mathbf{Z}_k, C)\right]
        && \text{(Factor out independent expectations.)}
    \end{aligned}
    \\
    &=\begin{aligned}[t]
        &\nabla_{\Theta_k}\mathbb{E}_{\prod_{n} q(\bm{c}_n)}\left[ \ln\left[\sum_{G_k}\nolimits  p(G_k\mid \mathbf{Z}_k) p(\Theta_k, \mathcal{D}\mid G_k, C)\right]\right]
        && \text{(From generative model.)} 
    \end{aligned}
    \\
    &=\begin{aligned}[t]
        &\mathbb{E}_{\prod_{n} q(\bm{c}_n)}\left[\nabla_{\Theta_k} \ln\left[ \mathbb{E}_{ p(G_k\mid \mathbf{Z}_k)}\left[ p(\Theta_k, \mathcal{D}\mid G_k, C)\right]\right]\right]
    \end{aligned}
    \\
    &=\begin{aligned}[t]
        &\mathbb{E}_{\prod_{n} q(\bm{c}_n)}\left[\nabla_{\Theta_k} \ln\left[ \mathbb{E}_{ p(G_k\mid\mathbf{Z}_k)}\left[p(\Theta_k\mid G_k) \prod_n\nolimits p(\mathbf{x}_n\mid G_k, \Theta_k)^{c_{nk}}\right]\right]\right]
    \end{aligned}
    \\
    &=\begin{aligned}[t]
        &\mathbb{E}_{\prod_{n} q(\bm{c}_n)}\left[\frac{\nabla_{\Theta_k} \mathbb{E}_{ p(G_k\mid\mathbf{Z}_k)}\left[p(\Theta_k\mid G_k) \prod_n\nolimits p(\mathbf{x}_n\mid G_k, \Theta_k)^{c_{nk}}\right]}{ \mathbb{E}_{ p(G_k\mid \mathbf{Z}_k)}\left[p(\Theta_k\mid G_k) \prod_n\nolimits p(\mathbf{x}_n\mid G_k, \Theta_k)^{c_{nk}}\right]}\right]
        && \text{($\nabla \ln f(x) = \frac{\nabla f(x)}{f(x)}$.)}
    \end{aligned}
    \\
    &=\begin{aligned}[t]
        &\mathbb{E}_{\prod_{n} q(\bm{c}_n)}\left[\frac{\mathbb{E}_{p(G_k\mid \mathbf{Z}_k)}\left[\nabla_{\Theta_k}\left[p(\Theta_k\mid G_k) \prod_n\nolimits p(\mathbf{x}_n\mid G_k, \Theta_k)^{c_{nk}}\right]\right]}{ \mathbb{E}_{ p(G_k\mid \mathbf{Z}_k)}\left[p(\Theta_k\mid G_k) \prod_n\nolimits p(\mathbf{x}_n\mid G_k, \Theta_k)^{c_{nk}}\right]}\right]
    \end{aligned}
\end{align*}
\endgroup
With samples $\{\bm{c}_n'\}_{n=1}^N, s=1,\ldots,S$ drawn from the variational distribution $ q(C)$, we can compute a Monte Carlo estimate of the gradient:
\begin{equation}
    \nabla_{\Theta_k}\ln q(\mathbf{Z}_k, \Theta_k) \approx \frac{1}{S}\sum_s \frac{\mathbb{E}_{p(G_k\mid \mathbf{Z}_k)}\left[\nabla_{\Theta_k}\left[p(\Theta_k\mid G_k) \prod_n\nolimits p(\mathbf{x}_n\mid  G_k, \Theta_k)^{c_{nk}^{(s)}}\right]\right]}{\mathbb{E}_{p(G_k\mid \mathbf{Z}_k)}\left[p(\Theta_k\mid G_k) \prod_n\nolimits p(\mathbf{x}_n\mid  G_k, \Theta_k)^{c_{nk}^{(s)}}\right]}.
     \label{eq:vamsl_theta_approx_gradient}
\end{equation}

\textbf{Variational Responsibilities} The variational responsibilities are obtained by:
\begingroup
\allowdisplaybreaks 
\addtolength{\jot}{0.3em} 
\begin{align*}
    &\ln q^*(\bm{c}_n) \\
    &= \mathbb{E}_{ q(G,\Theta) q(\bm{\pi})\prod_{i \neq n} q(\bm{c}_i)}\left[\ln p(G,\Theta, C, \bm{\pi}, \mathcal{D})\right]
    \\
    &=\begin{aligned}[t]
        &\mathbb{E}_{ q(G,\Theta) q(\bm{\pi})\prod_{i \neq n} q(\bm{c}_i)}\left[\ln p(\mathcal{D}\mid G, \Theta, C))\right]
        && \text{(Expand generative model.)}
        \\
        &+ \mathbb{E}_{ q(G,\Theta) q(\bm{\pi})\prod_{i \neq n} q(\bm{c}_i)}\left[\ln p(\Theta \mid G)\right]
        \\
        &+ \mathbb{E}_{ q(G,\Theta) q(\bm{\pi})\prod_{i \neq n} q(\bm{c}_i)}\left[\ln p(G)\right]
        \\
        &+ \mathbb{E}_{ q(G,\Theta) q(\bm{\pi})\prod_{i \neq n} q(\bm{c}_i)}\left[\ln p(C\mid  \bm{\pi})\right]
        \\
        &+ \mathbb{E}_{ q(G,\Theta) q(\bm{\pi})\prod_{i \neq n} q(\bm{c}_i)}\left[\ln p(\bm{\pi})\right] 
    \end{aligned}
    \\
    &=\begin{aligned}[t]
        &\mathbb{E}_{ q(G,\Theta) q(\bm{\pi})\prod_{i \neq n} q(\bm{c}_i)}\left[\ln p(\mathcal{D}\mid G, \Theta, C))\right] 
        && \text{(Terms independent of $q(\bm{c}_n)$.)}
        \\
        &+ \mathbb{E}_{ q(G,\Theta) q(\bm{\pi})\prod_{i \neq n} q(\bm{c}_i)}\left[\ln p(C\mid \bm{\pi})\right] + \textrm{constant}
    \end{aligned}
    \\
    &=\begin{aligned}[t]
        &\mathbb{E}_{\prod_k\nolimits q(G_k,\Theta_k) q(\bm{\pi})\prod_{i \neq n} q(\bm{c}_i)}\left[\sum_j\nolimits \sum_k\nolimits c_{jk}\ln p(x_j\mid G_k, \Theta_k)\right]
        \\
        &+ \mathbb{E}_{\prod_k\nolimits  q(G_k,\Theta_k) q(\bm{\pi})\prod_{i \neq n} q(\bm{c}_i)}\left[\sum_j\nolimits \sum_k\nolimits c_{jk}\ln\pi_k\right] + \textrm{constant}  
    \end{aligned}
    \\
    &=\begin{aligned}[t]
        &\mathbb{E}_{\prod_k\nolimits  q(G_k,\Theta_k)}\left[\sum_k\nolimits c_{nk}\ln p(\mathbf{x}_n\mid G_k, \Theta_k)\right]
        && \text{(Terms where $j \neq n$ are constant and }
        \\
        &+ \mathbb{E}_{ q(\bm{\pi})}\left[\sum_k\nolimits c_{nk}\ln\pi_k\right] + \textrm{constant}
        && \text{(Expectations independent of the terms can be factored out.)}
    \end{aligned}
    \\
    &=\begin{aligned}[t]
        &\sum_k\nolimits c_{nk} \left[\mathbb{E}_{ q(G_k,\Theta_k)}\left[\ln p(\mathbf{x}_n\mid G_k, \Theta_k)\right] +  \mathbb{E}_{ q(\bm{\pi})}\left[\ln\pi_k\right]\right]
        + \textrm{constant}
    \end{aligned} 
\end{align*}
\endgroup

The above equality implies the proportionality:
\begin{equation*}
     q^*(\bm{c}_n) \propto \prod_k\nolimits \exp\left[\mathbb{E}_{ q(G_k,\Theta_k)}\left[\ln p(\mathbf{x}_n\mid G_k, \Theta_k)\right] +  \mathbb{E}_{ q(\bm{\pi})}\left[\ln\pi_k\right]\right]^{c_{nk}}.
\end{equation*}

From which the marginal  assignment probabilities are computed by normalizing across the components using the softmax function, such that:
\begin{equation*}
     q^*(c_{nk} = 1) = \gamma(c_{nk}) = \frac{\exp\left[\mathbb{E}_{ q(G_k,\Theta_k)}\left[\ln p(\mathbf{x}_n\mid G_k, \Theta_k)\right] +  \mathbb{E}_{ q(\bm{\pi})}\left[\ln\pi_k\right]\right]}{\sum_{h=1}^K \exp\left[\mathbb{E}_{ q(G_h,\Theta_h)}\left[\ln p(\mathbf{x}_n\mid G_h, \Theta_h)\right] +  \mathbb{E}_{ q(\bm{\pi})}\left[\ln\pi_h\right]\right]}.
\end{equation*}

\textbf{Complete Derivation of Mixing Weight Updates} The mixing weights are updated by:
\begingroup
\allowdisplaybreaks 
\addtolength{\jot}{0.3em} 
\begin{align*}
    &\ln q^*(\bm{\pi}) \\
    &= \mathbb{E}_{ q(G,\Theta) q(C)}\left[\ln p(G,\Theta, C, \bm{\pi}, \mathcal{D})\right]
    \\
    &=\begin{aligned}[t]
        &\mathbb{E}_{ q(G,\Theta) q(C)}\left[\ln p(\Theta, \mathcal{D}\mid G, C))\right] 
        && \text{(Expand generative model.)}
        \\
        &+ \mathbb{E}_{ q(G,\Theta) q(C)}\left[\ln p(G)\right]
        \\
        &+ \mathbb{E}_{ q(G,\Theta) q(C)}\left[\ln p(C\mid \bm{\pi})\right]
        \\
        &+ \mathbb{E}_{ q(G,\Theta) q(C)}\left[\ln p(\bm{\pi})\right] 
    \end{aligned}
    \\
    &=\begin{aligned}[t]
        &+ \mathbb{E}_{ q(G,\Theta) q(C)}\left[\ln p(C\mid \bm{\pi})\right]
        && \text{(Terms independent of $\bm{\pi}$ are constant w.r.t. to $ q(\bm{\pi})$.)}
        \\
        &+ \mathbb{E}_{ q(G,\Theta) q(C)}\left[\ln p(\bm{\pi})\right]  + \textrm{constant}
    \end{aligned}
    \\
    &=\begin{aligned}[t]
        &+ \mathbb{E}_{ q(C)}\left[\ln p(C\mid \bm{\pi})\right]
        && \text{(Expectations independent of the terms can be factored out.)}
        \\
        &+ \ln p(\bm{\pi}) + \textrm{constant}
    \end{aligned}
    \\
    &=\begin{aligned}[t]
        &+ \mathbb{E}_{\prod_i\nolimits q(\bm{c}_i)}\left[\sum_n\nolimits \sum_k\nolimits c_{nk}\ln\pi_k \right]
        && \text{($p(\bm{\pi}) = \textrm{Dir}(\bm{\alpha})$.)}
        \\
        &+ \ln\frac{1}{B(\bm{\alpha})} +  \sum_k (\alpha_k - 1)\ln\pi_k + \textrm{constant}
    \end{aligned}
    \\
    &=\begin{aligned}[t]
        &+ \sum_n\nolimits \sum_k\nolimits \mathbb{E}_{ q(c_{n})}\left[c_{nk}\ln\pi_k \right]
        && \text{(Terms where $i \neq n$ and $\ln\frac{1}{B(\bm{\alpha})}$ are constant w.r.t. $ q(\bm{\pi})$.)}
        \\
        &+ \sum_k (\alpha_k - 1)\ln\pi_k + \textrm{constant}
    \end{aligned}
    \\
    &=\begin{aligned}[t]
        &\sum_k\nolimits \ln\pi_k\left[(\alpha_k - 1) + \sum_n\nolimits\mathbb{E}_{ q(c_{n})}[c_{nk}]\right] + \textrm{constant}
    \end{aligned}
    \\
    &=\begin{aligned}[t]
        &\sum_k\nolimits \ln\pi_k\left[(\alpha_k + \sum_n\nolimits\gamma(c_{nk})) - 1\right] + \textrm{constant}
        && \text{(As $\ \mathbb{E}_{ q(c_{n})}[c_{nk}]=q^*(c_{nk} = 1) = \gamma(c_{nk})$.)}
    \end{aligned}
\end{align*}
\endgroup

\noindent Showing the update is a Dirichlet distribution $ q^*(\bm{\pi}) = \textrm{Dir}(\bm{\alpha}^*)$, where $ \alpha_k^* = \alpha_k + \sum_n\nolimits\gamma(c_{nk})$.

\subsubsection{Derivation of Proposition~\ref{prop:LMPE}}
\label{app:latent_mixture_posterior_expectation_derivation}
Below we derive the latent mixture posterior expectation which connects the expectation of a function with respect to a joint distribution over graphs $G_k$ and parameters $\Theta_k$ with an expectation with respect to a distribution over latent graph embeddings $\mathbf{Z}_k$, parameters $\Theta_k$, and assignments $C$.
\begingroup
\allowdisplaybreaks 
\addtolength{\jot}{0.3em} 
\begin{align*}
    &\mathbb{E}_{p(G_k,\Theta_k \mid \mathcal{D})}[f(G_k,\Theta_k)] \\
    &= \int_{\Theta_k}\sum_{G_k} p(G_k,\Theta_k \mid \mathcal{D})f(G_k,\Theta_k) \dd \Theta_k \\
    &= \int_{\Theta_k}\sum_{G_k} \sum_{C} p(G_k,\Theta_k, C \mid \mathcal{D})f(G_k,\Theta_k) \dd \Theta_k \\
    &= \int_{\Theta_k}\sum_{G_k} \sum_{C} \frac{p(C)p(G_k)p(\Theta_k\mid G_k)p(\mathcal{D}\mid G_k,\Theta_k, C)}{p(\mathcal{D})}f(G_k,\Theta_k) \dd \Theta_k\\
    &= \int_{\Theta_k}\sum_{G_k} \sum_{C} \int_{\mathbf{Z}_k} \frac{p(C)p(\mathbf{Z}_k)p(G_k\mid \mathbf{Z}_k)p(\Theta_k\mid G_k)p(\mathcal{D}\mid G_k,\Theta_k, C)}{p(\mathcal{D})}f(G_k,\Theta_k) \dd \Theta_k \dd \mathbf{Z}_k \\
    \intertext{\hspace{2cm} as \ $ p(\mathbf{Z}_k,\Theta_k, C \mid \mathcal{D}) = \frac{p(\Theta_k, \mathcal{D}\mid \mathbf{Z}_k, C)p(C)p(\mathbf{Z}_k)}{p(\mathcal{D})} \iff \frac{p(C)p(\mathbf{Z}_k)}{p(\mathcal{D})} = \frac{p(\mathbf{Z}_k,\Theta_k, C \mid \mathcal{D})}{p(\Theta_k, \mathcal{D}\mid \mathbf{Z}_k, C)}$,}
    &= \int_{\Theta_k}\sum_{G_k} \sum_{C} \int_{\mathbf{Z}_k} \frac{p(\mathbf{Z}_k,\Theta_k, C \mid \mathcal{D})p(G_k\mid \mathbf{Z}_k)p(\Theta_k\mid G_k)p(\mathcal{D}\mid G_k,\Theta_k, C)}{p(\Theta_k, \mathcal{D}\mid \mathbf{Z}_k, C)}f(G_k,\Theta_k) \dd \Theta_k \dd \mathbf{Z}_k\\
    &= \int_{\mathbf{Z}_k} \int_{\Theta_k} \sum_{C} p(\mathbf{Z}_k,\Theta_k, C \mid \mathcal{D}) \frac{\sum_{G_k} p(G_k\mid \mathbf{Z}_k)p(\Theta_k\mid G_k)p(\mathcal{D}\mid G_k,\Theta_k, C)f(G_k,\Theta_k)}{p(\Theta_k, \mathcal{D}\mid \mathbf{Z}_k, C)} \dd \Theta_k \dd \mathbf{Z}_k\\
    &= \int_{\mathbf{Z}_k} \int_{\Theta_k} \sum_{C} p(\mathbf{Z}_k,\Theta_k, C \mid \mathcal{D}) \frac{\sum_{G_k} p(G_k\mid \mathbf{Z}_k)p(\Theta_k\mid G_k)p(\mathcal{D}\mid G_k,\Theta_k, C)f(G_k,\Theta_k)}{\sum_{G_k} p(G_k\mid \mathbf{Z}_k) p(\Theta_k, \mathcal{D}\mid G_k, C)} \dd \Theta_k \dd \mathbf{Z}_k\\
    &= \int_{\mathbf{Z}_k} \int_{\Theta_k} \sum_{C} p(\mathbf{Z}_k,\Theta_k, C \mid \mathcal{D}) \frac{\sum_{G_k} p(G_k\mid \mathbf{Z}_k)p(\Theta_k\mid G_k)p(\mathcal{D}\mid G_k,\Theta_k, C)f(G_k,\Theta_k)}{\sum_{G_k} p(G_k\mid \mathbf{Z}_k) p(\Theta_k\mid G_k) p(\mathcal{D}\mid G_k, C)} \dd \mathbf{Z}_k \dd \Theta_k \\
    &= \int_{\mathbf{Z}_k} \int_{\Theta_k} \sum_{C} p(\mathbf{Z}_k,\Theta_k, C \mid \mathcal{D}) \frac{\mathbb{E}_{p(G_k\mid \mathbf{Z}_k)}\left[p(\Theta_k\mid G_k)p(\mathcal{D}\mid G_k,\Theta_k, C)f(G_k,\Theta_k)\right]}{\mathbb{E}_{p(G_k\mid \mathbf{Z}_k)} \left[ p(\Theta_k\mid G_k) p(\mathcal{D}\mid G_k, C)\right]} \dd \mathbf{Z}_k \dd \Theta_k \\
    &= \mathbb{E}_{p(\mathbf{Z}_k,\Theta_k, C \mid \mathcal{D})}\left[\frac{\mathbb{E}_{p(G_k \mid \mathbf{Z}_k)}[f(G_k,\Theta_k)p(\Theta_k \mid G_k)\prod_n p(\mathbf{x}_n\mid G_k,\Theta_k)^{c_{nk}}] }{\mathbb{E}_{p(G_k\mid \mathbf{Z}_k)}[p(\Theta_k \mid G_k)\prod_n p(\mathbf{x}_n\mid G_k,\Theta_k)^{c_{nk}}] }\right].
\end{align*}
\endgroup

\section{USER MODEL FOR IMAGINARY OBSERVATIONS}
\label{app:user_model}

We employ the device of \textit{imaginary observations} \citep{consonniPriorDistributionsObjective2018, goodProbabilityWeighingEvidence1950a} to formalize the background knowledge $\mathcal{K}_{ij}$ an expert has about the existence of a causal effect $i\to j$ as a set of idealized trials $\mathcal{K}_{ij} = (k, n)$ confirming (refuting) said effect. Starting from a beta distributed prior belief $p(\psi_{ij})=\textrm{Beta}(\psi_{ij}\mid \alpha_0, \beta_0)$ with the mode $\psi_{ij,0}^{*}=\frac{ \alpha_0-1}{ \alpha_0+ \beta_0-2}$, where $\alpha_0,\beta_0>1$ for the mode to be defined, the expert's edge belief is formalized as the mode of the posterior belief $p(\psi_{ij}\mid \mathcal{K}_{ij}) \propto p(\mathcal{K}_{ij}\mid \psi_{ij})p(\psi_{ij})$, conditioned on the aforementioned trials.

Due to our assumed beta-binomial conjugate model and imaginary observations from deterministic and correct experiments for confirming (refuting) causal effects, we know the mode of the posterior $\psi_{ij}^*$ (expressed by the expert) must have moved in the direction supported by the observations. We can therefore derive a mapping from the posterior mode to imaginary observations by rearranging the conjugate update of the posterior mode, relating the posterior mode to the prior mode as well as the number of confirming and disconfirming trials:
\begin{equation*}
    \psi_{ij}^* = \frac{\alpha_0 + k-1}{\alpha_0 + k + \beta_0 + (n-k)-2}.
\end{equation*}
When the posterior mode has increased relative to the prior mode $\psi_{ij}^* > \psi_{ij,0}^*$, the trials are confirmatory and we can conclude $k=n$. While the opposite case $\psi_{ij}^* < \psi_{ij,0}^*$ indicates the trials disconfirm the existence of a causal effect and hence $k=0$. Solving for these case we get the following mapping from expert belief to imaginary observations:
\begin{equation}
    f_{\alpha_0, \beta_0}(\psi_{ij}^*)
    =\begin{cases}
        \mathcal{K}_{ij} =  (n_{ij}=\left\lfloor\frac{\psi_{ij}^*(\alpha_0+\beta_0-2)-\alpha_0+1}{1-\psi_{ij}^*}\right\rfloor, k_{ij}=n_{ij}),
        \quad&\textrm{if }\ \psi_{ij}^* > \psi_{ij,0}^{*};\\[4pt]
        \mathcal{K}_{ij} = (n_{ij} =\left\lfloor\frac{\alpha_0-1-\psi_{ij}^*(\alpha_0+\beta_0-2)}{\psi_{ij}^*}\right\rfloor, k_{ij}=0),
        \quad&\textrm{if }\  \psi_{ij}^* < \psi_{ij,0}^{*}. 
    \end{cases} \notag
\end{equation}
Note that, as the prior parameters $\alpha_0,\beta_0$ determine how many observations it would take for the prior mode to shift to the mode expressed by the expert, they effectively determine the informativeness of the informative graph prior. With low values constituting a low informativeness and high values making the prior very informative. In our experiments we use the same prior parameters for all edges, if desired these could vary over the edges.

\section{IMPLEMENTATION OF INFERENCE FOR THE USER MODEL}\label{app:user_model_implementation}
The construction described in Section~\ref{sec:informative_elicitation_graph_prior} yields an elicitation prior that deterministically drives particles in the direction of edge inclusion or exclusion that the expert has endorsed. This is comparable to the method proposed by \citet{geigerLearningGaussianNetworks1994}, which scores networks based on their distance from a reference prior network. However, to include stochasticity in a sample of particles from the posterior we propose sampling separate subsets of the elicited responses, termed elicitation matrices, for each particle. For each particle $\mathbf{Z}^{(p)}$ to be optimized, we sample an elicitation matrix $E^{(p)} \mid \mathcal{D}_S$, by independently sampling its entries $E_{ij}^{(p)}$ each of which takes one of two possible values: $\psi_{ij}^{*}$ and $0.5$, with $P(E_{ij}^{(p)} = \psi_{ij}^*)=\max(\psi_{ij}^*,1-\psi_{ij}^*)$. 
In other words, the closer a response from the expert is to $0$ or $1$ (i.e., the stronger the opinion), the more probable it is to be part of a particle's elicitation matrix. Whereas weakly opinionated responses (i.e., close to $0.5$) are more probable to not be included (i.e., labeled ``unresponded'' and $\equiv0.5$). Rather than deterministically biasing the inference towards the expert opinion, this leads the proportion of particles in the particle distribution with priors to (exclude) include an edge to correspond to the probability the expert has responded with. For a gradient based method of inference, such as the method proposed in \Cref{sec:Method}, the elicited prior gradient for each particle is therefore:
\begin{align*}
    \nabla_{\mathbf{Z}} \ln p(\mathbf{Z}^{(p)} \mid f(E^{(p)})) = \nabla_{ \mathbf{Z}}\ln p(\mathbf{Z}^{(p)} \mid \mathcal{D}^{(p)}_T) = \nabla_{ \mathbf{Z}}\ln p(\mathbf{Z}^{(p)}) +  \nabla_{\mathbf{Z}}\ln p(\mathcal{D}^{(p)}_T \mid \mathbf{Z}^{(p)}),
\end{align*}
which is straightforward to add to the SVGD implementation of DiBS provided by \citet{lorchDiBSDifferentiableBayesian2021a}.

\section{APPROXIMATING THE EXPECTED INFORMATION GAIN}
\label{app:approximating_EIG}

Bayesian experimental design \citep{rainforthModernBayesianExperimental2024} defines a method for choosing optimal experiments (with respect to a given utility) under uncertainty. Given a parameter distribution $p(\mathbf{Z} \mid \mathcal{D})$ and a simulator $p(\psi_{ij}^*\mid \mathbf{Z}, \xi_{ij})$, we apply the BED framework to pick optimal queries about causal edges for the expert based on the information theoretic utility of each query. The optimal query maximizes the expected information gain, which can be expressed as:
\begin{equation*}
    \underset{\xi_{ij} \in \Xi_k}{\textrm{argmax}}\hspace{1mm} \textrm{EIG}(\xi_{ij})  
    = \underset{\xi_{ij} \in \Xi}{\textrm{argmax}}\hspace{1mm}                           
        \mathbb{E}_{p(\mathbf{Z}\mid \mathcal{D})p(\psi_{ij}^*\mid \mathbf{Z}, \xi_{ij})} 
        \left[\ln p(\psi_{ij}^* \mid \mathbf{Z}, \xi_{ij}) - \ln p(\psi_{ij}^* \mid \xi_{ij})\right]. 
\end{equation*}
However, computing the EIG is intractable and we instead need to approximate it. As the domain of our response variables $\psi_{ij}^* \in [0,1]$ is continuous, we propose using the nested Monte Carlo estimator~(NMC) $\hat{\mu}_{NMC}$ which can be computed by sampling from our parameter distribution and simulator. Using samples $\mathbf{Z}^{(p)} \sim p(\mathbf{Z} \mid \mathcal{D}),\ p=1,\ldots,P$ and  $ \psi_{ij}^{*(s)}, \mathbf{Z}^{(s)} \sim p(\psi_{ij}^*\mid \mathbf{Z}, \xi_{ij})p(\mathbf{Z} \mid \mathcal{D}),\ s=1,\ldots,S$, the NMC-estimate can be computed as:
\begin{equation}\label{eq:NMC_estimator} 
    \hat{\mu}_{NMC} = \frac{1}{S}\sum_{s=1}^{S}\ln \frac{p(\psi_{ij}^{*(s)}\mid \mathbf{Z}^{(s)}, \xi_{ij})}
                                            {\frac{1}{P}\sum_{p=1}^P p(\psi_{ij}^{*(s)}\mid \mathbf{Z}^{(p)}, \xi_{ij})}.
\end{equation}
We note that if the expert only needed to respond with hard constraints (e.g., $\psi_{ij}^* \in \{0,1\}$), it would be possible to enumerate the expectation over $p(\psi_{ij}^*\mid \mathbf{Z}, \xi_{ij})$ in the EIG. It would then be possible to instead approximate the EIG using a Rao-Blackwellized estimator and a Bernoulli distributed simulator. We refer the reader to \citet{rainforthModernBayesianExperimental2024} for further details on BED and approximating the EIG.

\section{ADDITIONAL DETAILS OF NUMERICAL AND BREAST CANCER EXPERIMENTS}\label{app:numerical-experiments}

\subsection{Metrics Used for Experiments}\label{app:metrics}
The structured Hamming distance (SHD) measures how different two graphs are. This distance is given by counting the number of incorrect edges, meaning the number of changes (flips, additions or removals) of edges to move from one of the graphs to the other.

For (out-of-sample) classification accuracy and predictive performance in the synthetic case we generate a held-out dataset $\mathcal{D}_{ho} = \{\tilde{\mathbf{x}}_{n} \mid \tilde{\mathbf{x}}_n\in \mathbb{R}^{20}\}_{n=1}^{100}$ generated from a single ground truth BN in the homogeneous setting (\Cref{sec:results_homogeneous}) and, in the heterogeneous setting (\Cref{sec:hetergenous_experiments}), from two BNs with mixing probabilities $(0.5,0.5)$. In the breast cancer labeling assignment (\Cref{sec:application}) we use $10$-fold cross validation. To compute the classification accuracy and predictive performance we use the component ordering of the components that maximizes the (in-sample) MAP classification accuracy using the modified Jonker-Volgenant algorithm\footnote{implemented in Python SciPy \citep{virtanenSciPy10Fundamental2020}.} described by \citet{crouseImplementing2DRectangular2016}. The ESHD for heterogeneous experiments is the sum of the ESHDs for each component with respect to the ground truth graph indicated by the ordering given by maximizing the in-sample MAP classification accuracy. 

We assess the predictive performance using the MAP classifications of out-of-sample observations to components (denoted $\tilde{k}(n) = \underset{k}{\textrm{argmax}} \ q(\tilde{c}_{nk}=1)$) and computing the \textit{(MAP)} \textit{negative log point-wise predictive density} \citep{gelmanUnderstandingPredictiveInformation2014}:
\begin{align*}
    &\textrm{MAP Neg.}\ \mathcal{LPPD}(\mathcal{D}_{ho}) =\ \sum_{k}\sum_n \mathbf{1}_{\{k=\tilde{k}(n)\}}\log\mathbb{E}_{p(G_{k}, \Theta_{k}\mid \mathcal{D})}\left[p(\tilde{\mathbf{x}}_{n}\mid G_{k}, \Theta_{k})\right].
\end{align*}

\subsection{Settings for Numerical Experiments}\label{app:num-settings}
We instantiate VaMSL with Gaussian BNs parametrized by linear or non-linear mean functions and observational noise set to $\sigma^2=0.1$. Where non-linearity is instantiated using 2-layer neural networks (NN) with 5 neurons per hidden layer and ReLU activation functions. All the parameters for the NNs and linear models are randomly set with a standard normal distribution. For all the experiments we take a total 6000 gradient steps for SVGD. The step size $\eta_t$ is set adaptively using RMSProp\footnote{find an implementation in the Python JAX package \citep{jax2018github} at \href{http://github.com/jax-ml/jax}{http://github.com/jax-ml/jax}.} with a learning rate of $0.005$. In the mixture case we do 10 CAVI updates, iteratively updating first the SVGD particle distribution and, subsequently, the responsibilities and the mixing weight distribution. For the first 9 CAVI updates the SVGD particle optimization consist of 50 gradient steps while the last one consists of the final 5500 steps. The annealing schedules for the temperatures are set such that for step $t$ we use $\beta_t = t$ as well as $\omega_t = 0.2t$ for linear BNs and $\omega_t = 0.02t$ for non-linear BNs. For the length-scales of the SVGD kernel, given in \Cref{eq:additive_SE_kernel}, we use $\gamma_Z=5$ as well as $\gamma_\Theta=500$ for linear BNs and $\gamma_\Theta=1000$ for non-linear BNs.  When updating the responsibilities, for the first 9 CAVI updates, we approximate the expectation over $q(G_k, \Theta_k)$ in \Cref{eq:variational_C_update_marginal} using latent graphs embeddings, as enabled by \Cref{prop:LMPE}, and use all particles available. As explained in \Cref{sec:vamsl-algorithm}, it might be necessary to randomly restart VaMSL for optimal performance, in our experiments we use a maximum of 5 random restarts. 

For the querying strategy we use 60 particles for inference, for the mixture of graphs we use 60 particles per component for VaMSL and 120 particles for DiBS. In the cancer labeling task we use DiBS with 120 particles to infer the MAP graphs used as expert information and we use 15 particles per component when inferring the mixture model. For the observation noise For all experiments, we set $\ell=d$. To compute the approximate gradients for VaMSL, given in \Cref{eq:vamsl_z_approx_gradient} and \Cref{eq:vamsl_theta_approx_gradient}, we use four Monte Carlo samples. Otherwise we use values reflecting the default values of the DiBS implementation\footnote{found at \href{https://github.com/larslorch/dibs}{https://github.com/larslorch/dibs}.} of \citet{lorchDiBSDifferentiableBayesian2021a}. We do not use their weighted particle mixture, termed DiBS+.

For the informative prior we set $\alpha_0=\beta_0=10$ in the synthetic experiments $\alpha_0=\beta_0=1000$ in the real-world experiments. When calculating the EIG of a query, through \Cref{eq:NMC_estimator}, we use all particles available.

For the GMM we use default values and initialize the parameters from randomly selected data points. For the causal k-means we use the implementation\footnote{found at \href{https://causal.dev/code/fibroblast_clustering.py}{https://causal.dev/code/fibroblast\_clustering.py}.} given by \cite{markhamDistanceCovariancebasedKernel2022}, using the default value ($0.1$) for the significance level. Experiments were run on NVIDIA H100 GPUs.

Figure~\ref{fig:VaMSL_intro_example} features inference with $100$ observations from a two component mixture, with mixing probabilities (0.5, 0.5), of linear Gaussian BNs with ER graphs over $4$ variables and an expectation of $1$ edge per node. VaMSL was initiated with one correct soft constraint per component about a random edge's existence. Otherwise all settings were the same as the default settings listed for synthetic data above.  

\subsection{Expert Oracle}\label{app:expert-simulation}
To assess the effect of incorporating expert information we use an expert oracle. In the synthetic case, the oracle's answers are based on the ground truth graph that was used to generate the synthetic data, while in the real-world data setting the answers are based on the MAP graphs obtained when running DiBS on the homogeneous populations of a subset of the breast cancer data not used in the presented experiments. For our real world experiments we used half of the data to infer the graphs for the expert oracle and the other half for benchmarking our method. 

The expert oracle returns probabilities given edge queries by sampling a value from a beta distribution parametrized by a mean and a variance parameter. The mean is determined by the reliability $r\in[0,1]$ of the expert and the binary value indicating the edge's existence in the corresponding expert graph. It is calculated according to $\left|r-1 + \mathbf{1}_{(i\to j)}\right|$. 
The variance is set to 0.05 and the default value for the reliability is $r=0.9$. 

\subsection{Additional Results of Erd\H{o}s-R\'enyi Numerical Experiments}\label{app:num-details-ER}
In this subsection we provide supplementary numerical results to the example shown in the main text. These results use an Erd\H{o}s-R\'enyi graph-prior, like the one in the main text, which corresponds to $p(G_\omega(\mathbf{Z}_k))\propto q^{\lVert G_{\omega}(\mathbf{Z}_k)\rVert_1}(1-q)^{\binom{d}{2}-\lVert G_{\omega}(\mathbf{Z}_k)\rVert_1}$, where $q$ is the probability for any given edge to exist.

\begin{figure}[!htp]
  \centering
  \includegraphics[width=\textwidth]{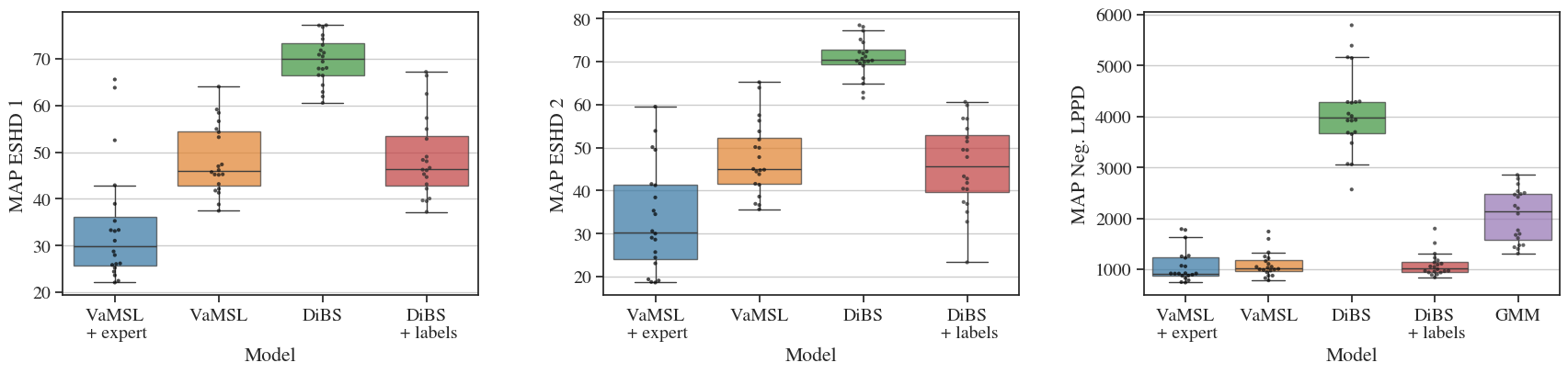}
      \caption{Boxplots and values of metrics in held-out observations, simulated with \textbf{linear} Gaussian ER BNs. \textit{Left}: ESHD in the first component, \textit{Center}:  ESHD in the second component, \textit{Right}: predictive power by the log point-wise predictive density. ESHD not available for Causal k-means and GMMs.}
      \label{fig:ER-linear-graph-metrics}
  \end{figure}

In Figure~\ref{fig:ER-linear-graph-metrics} we show the metrics presented in the main text and Appendix~\ref{app:metrics} for the linear Gaussian Erd\H{o}s R\'enyi graphs. These metrics are computed for out-of-sample data, trained using the same setting described in the main text in Section~\ref{sec:results} with a linear parametrization. The results show how VaMSL~+~expert performs much better than the competing baselines in terms of structure learning as assessed by ESHD (left panels). In terms of predictive power (top-right) assessed by the log-predictive density it performs about as good as the gold-standard case of having the labels (DiBS+labels). Similarly for classification accuracy (bottom-right), both VaMSL with and without expert perform just as good as having the labels, and perform much better than both classical DiBS and the Gaussian mixture model.

\begin{figure}[!htp]
      \centering
  \includegraphics[width=\textwidth]{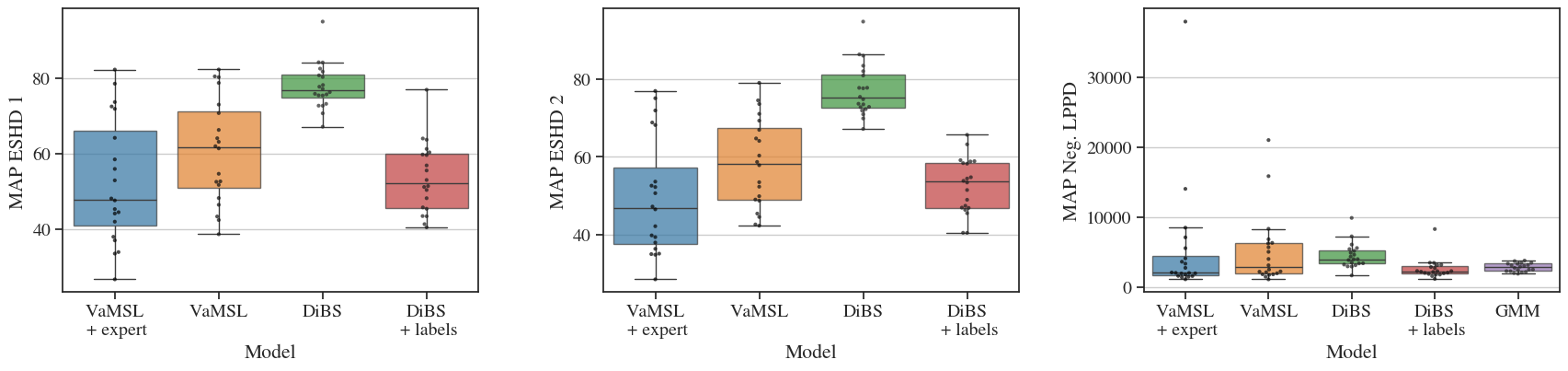}
  \caption{Boxplots of metrics of interest for simulations in \textbf{non-linear} Gaussian ER BNs. \textit{Top-left}: ESHD in the first component, \textit{Bottom-left}:  ESHD in the second component, \textit{Right-top}: predictive power by the log point-wise predictive density, \textit{Bottom-right}: percentage of correctly labeled samples. ESHD and neg-lppd not available for Causal k-means and GMMs.}
  \label{fig:ER-nonlinear-graph-metrics}
\end{figure}

Figure~\ref{fig:ER-nonlinear-graph-metrics} shows how in the non-linear case the same conclusion is reached, meaning that the type of parametrization is not affecting our conclusions. Structure learning is performing better, even than DiBS with labels---the gold standard case. Regarding the predictions, the log-predictive density shows a similar performance and even better than the GMM, with a better performance when the expert is included.

\clearpage
\subsection{Additional Results in Scale-free Numerical Experiments}\label{app:num-details-SF}
In this subsection we provide additional numerical results for the scale-free setting, where the density corresponds to $p(G_\omega(\mathbf{Z}_k))\propto \prod_{i=1}^d (1+\lVert G_{\omega}(\mathbf{Z}_i)\rVert_1)^{-3}$, where $G_\omega(\mathbf{Z})_i$ is the $i$-th row of $G_\omega(\mathbf{Z})$.

\begin{figure}[!htp]
  \centering
  \includegraphics[width=\textwidth]{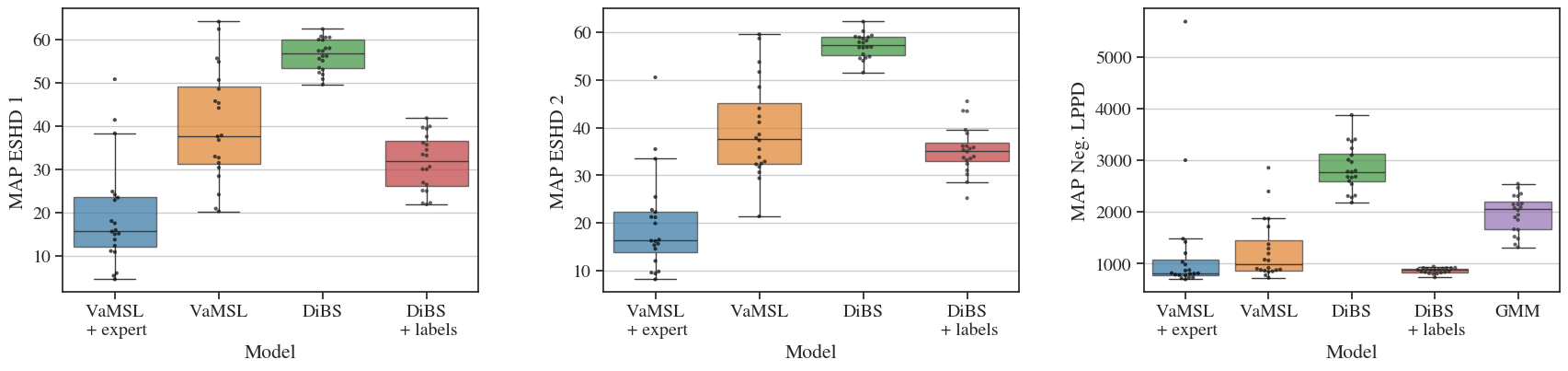}
      \caption{Boxplots of metrics of interest for simulations in \textbf{linear} Gaussian SF BNs. \textit{Left}: ESHD in the first component, \textit{Center}:  ESHD in the second component, \textit{Right}: predictive power by the neg. lppd. ESHD and neg-lppd not available for Causal k-means and GMMs. Neg. lppd not available for Causal k-means.}
      \label{fig:SF-linear-graph-metrics}
  \end{figure}

\begin{figure}[!htp]
      \centering
  \includegraphics[width=\textwidth]{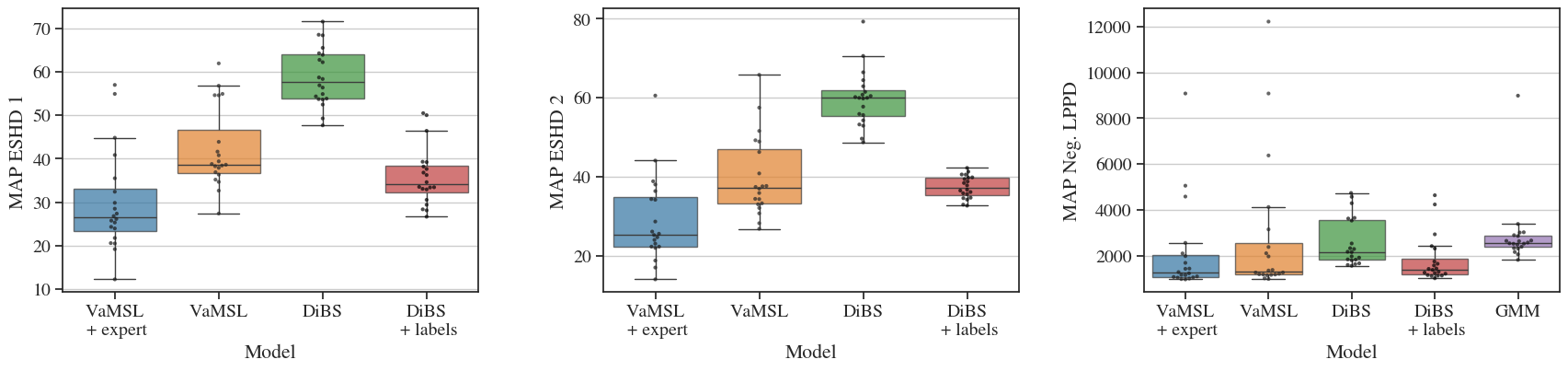}
  \caption{Boxplots of metrics of interest for simulations in \textbf{non-linear} Gaussian SF BNs. \textit{Left}: ESHD in the first component, \textit{Center}:  ESHD in the second component, \textit{Right}: predictive power by the neg. lppd. ESHD not available for Causal k-means and GMMs. Neg. lppd not available for Causal k-means.}
  \label{fig:SF-nonlinear-graph-metrics}
\end{figure}

Figures~\ref{fig:SF-linear-graph-metrics} and~\ref{fig:SF-nonlinear-graph-metrics} display how our proposed approach still performs better than the competing approaches. Our proposed method, VaMSL~+~expert outperforms all the other approaches in structure learning, and performs on par to the ideal scenario: classification when the labels are available at training time.

\begin{figure}[!htp]
  \centering
  \subfloat{\includegraphics[width=0.65\textwidth]{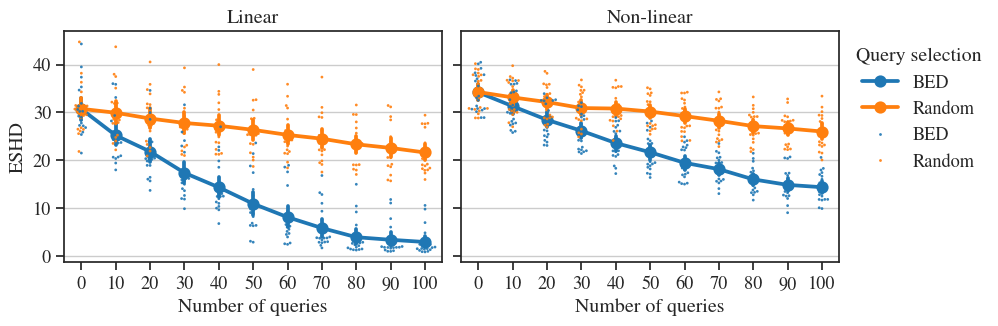}}
  \hfill
  \subfloat{\includegraphics[width=0.65\textwidth]{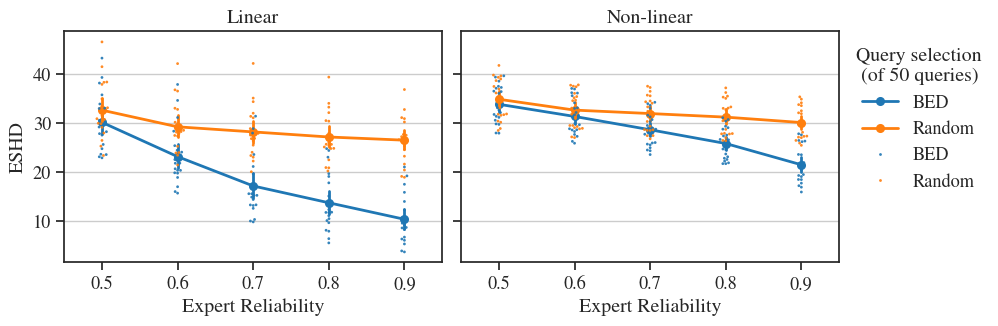}}
  \caption{Comparison between BED and random querying of the expert with varying number of queries (\textit{top}) and varying expert reliability (\textit{bottom}) when inferring Gaussian SF BNs, displayed by average performance with 95\% bootstrapped confidence interval, and the obtained ESHD. For both \textbf{linear} (\textit{left}) and \textbf{non-linear} graphs (\textit{right})}
  \label{fig:SF-graph-expert-sims}
\end{figure}

\clearpage
In Figure~\ref{fig:SF-graph-expert-sims} we display how our BED querying strategy has a marked improvement over the random approach, this applies to both possibilities considered, with varying number of queries (top) and with different levels of expert reliability (bottom), in both linear (left) and non-linear settings.
\subsection{Additional Results and Details of Breast Cancer Dataset}\label{app:cancer-details}
\begin{figure}[!htp]
    \centering
    \includegraphics[width=\linewidth]{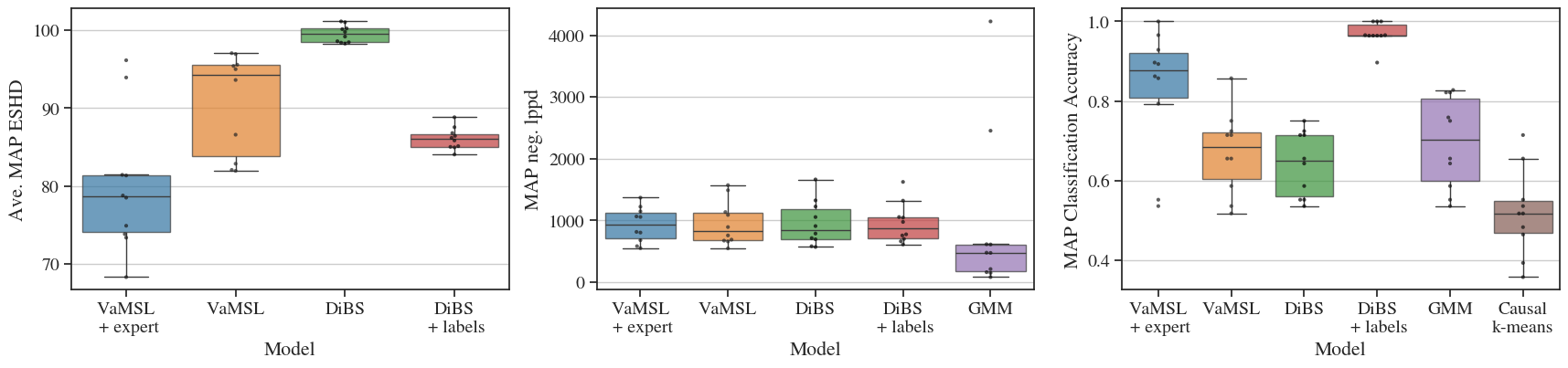}
    \caption{Boxplots and values for metrics of interest in out-of-sample data for breast cancer data set in methods of interest. \textit{Left}: average ESHD, \textit{Center}:  predictive power assessed by the log point-wise predictive density, \textit{Right}: percentage of correctly labeled samples. ESHD and neg. lppd not available for Causal k-means, and ESHD not available for GMMs.}
    \label{fig:UCI-metrics}
\end{figure}

To complement the main result in Section~\ref{sec:application}, in Figure~\ref{fig:UCI-metrics} we display other metrics of interest in the applied setting of the breast cancer dataset. Namely, the average MAP ESHD (left), the MAP neg. lppd (center), and the classification accuracy (right). These plots repeat the main message from the simulated experiments: VaMSL~+~expert performs much better in terms of structure learning and classification accuracy, while preserving a competitive predictive power.

\subsubsection{Expert Queries in Breast Cancer Dataset}\label{app:cancer-expert-queries}
\begin{figure}[!htp]
    \centering
    \includegraphics[width=\linewidth]{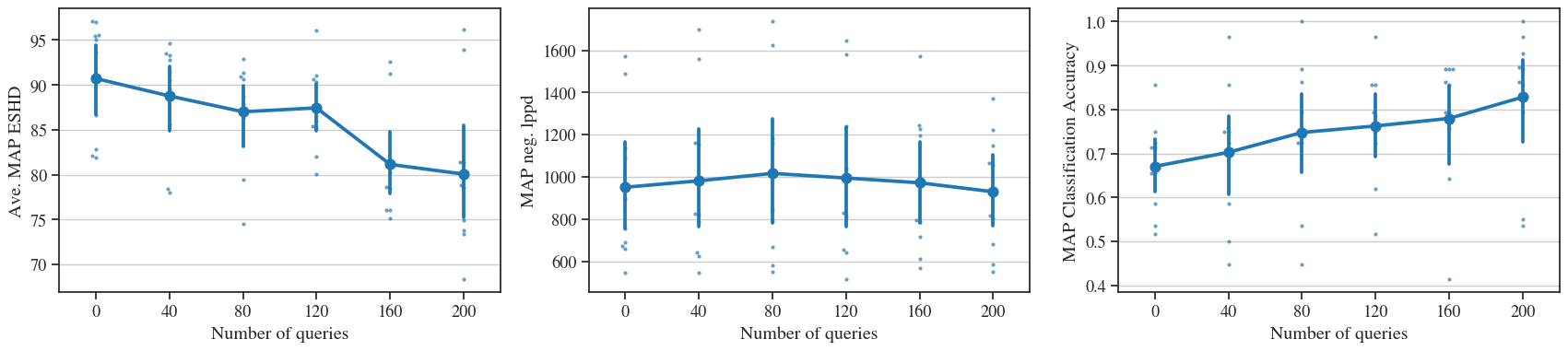}
    \caption{Mean and 95\% (bootstrapped) confidence interval of metrics for out-of-sample data using VaMSL with the BED querying strategy in the breast cancer dataset, modeled with SF priors and non-linear models. Assessed by (\textit{left}) average ESHD, (\textit{center}) MAP negative log-pointwise predictive density, and (\textit{right}) classification accuracy.}
    \label{fig:UCI-querying}
\end{figure}

In Figure~\ref{fig:UCI-querying} we showcase the performance of VaMSL with an increasing number of queries in the breast cancer dataset. The left panel shows how the graph distance diminishes with the number of queries, as is to be expected. Specifically, it  shows how the posterior graphs become closer to the assumed graph as the number of queries increases. The center panel shows how despite an increased performance in the graph distribution (left panel), the raw predictive power~---~as assessed by the MAP negative log-predictive density (see Appendix~\ref{app:metrics}) is not increasing. Nevertheless, the correct classification is still being obtained as shown in the right panel. We suspect that the vast dimensionality of the parameter space (since we are in non-linear models) and possible multi-modal topography of the loss function explains the lack of improved performance in the log-predictive density domain. However, our two main concerns do improve with an increased number of queries: obtaining the correct graph structure (ESHD), and obtaining an improved classification (classification accuracy).

\subsubsection{Breast Cancer Dataset Modeled with Erd\H{os}-R\'enyi Priors}
\begin{figure}
    \centering
    \includegraphics[width=\linewidth]{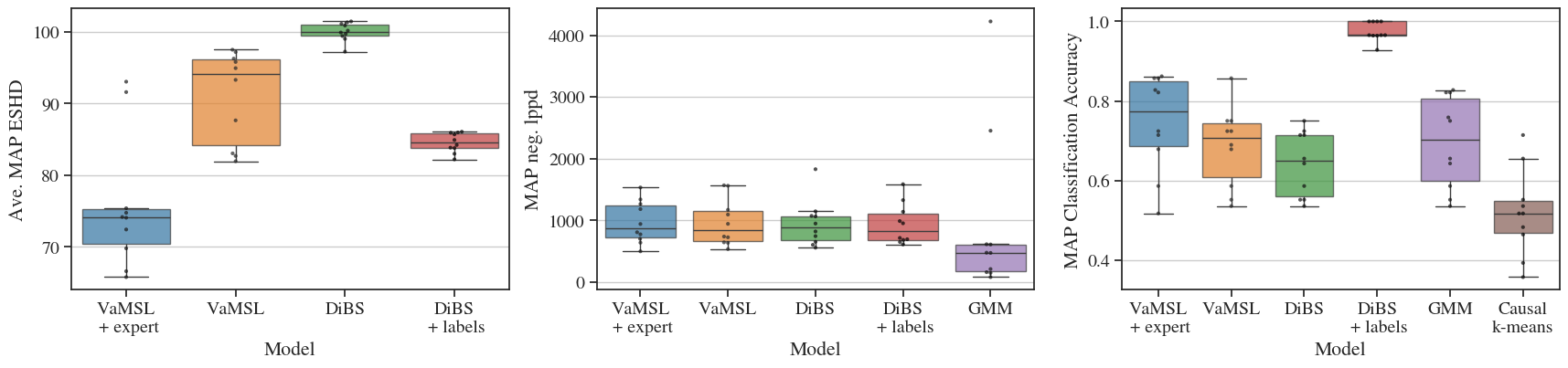}
    \caption{Boxplots and values for metrics of interest in out-of-sample data for breast cancer data set in methods of interest using ER priors. \textit{Left}: average ESHD, when compared to a perfect expert model, \textit{Center}:  predictive power assessed by the log point-wise predictive density, \textit{Right}: percentage of correctly labeled samples. ESHD and neg. lppd not available for Causal k-means, and ESHD not available for GMMs.}
    \label{fig:UCI_all_metrics_nonlinear_er}
\end{figure}

To confirm the performance of the model in the applied setting with the ER priors, we perform the same routines as in Section~\ref{sec:application}. In Figure~\ref{fig:UCI_all_metrics_nonlinear_er} we replicate Figure~\ref{fig:UCI-metrics} using ER priors instead of the SF priors used in the main text. The same conclusion holds, albeit at a smaller magnitude. The expert-informed model (VaMSL~+~expert) outperforms the other models in terms of classification (right panel), Additionally, the structure learning (left panel) is vastly better, and the predictive performance (center) is on-par to the other methods.

\begin{figure}[!htp]
    \centering
    \includegraphics[width=\linewidth]{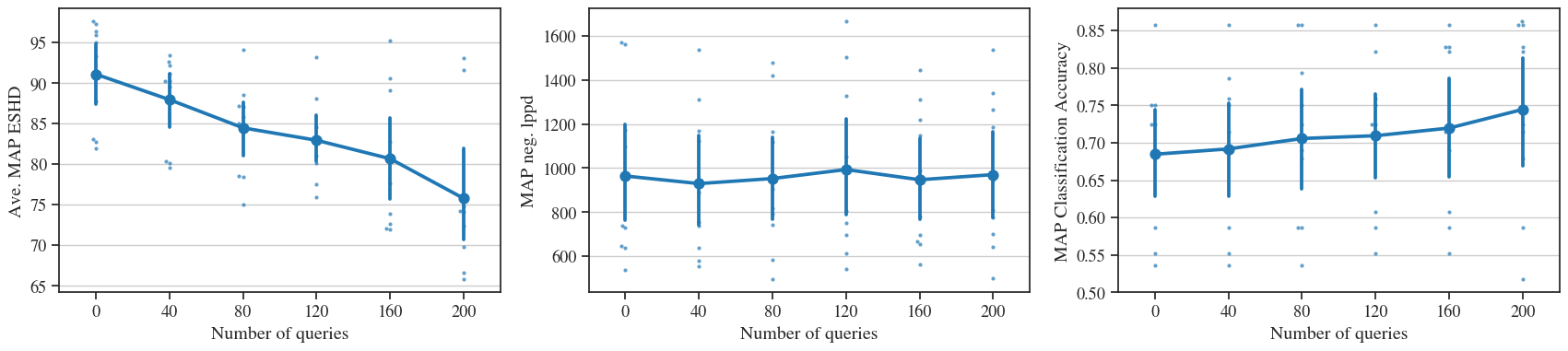}
    \caption{Mean and 95\% (bootstrapped) confidence interval of metrics for out-of-sample data using VaMSL with the BED querying strategy in the breast cancer dataset, modeled with ER priors and non-linear models. Assessed by (\textit{left}) average ESHD, (\textit{center}) MAP negative log-pointwise predictive density, and (\textit{right}) classification accuracy.}
    \label{fig:UCI-querying-ER}
\end{figure}

In Figure~\ref{fig:UCI-querying-ER} we show how additional queries to the expert do improve the method in the two main metrics of interest: labeling accuracy (right) and structure learning. As such, validating our query strategy. 

\subsection{Ablation Studies on Hyperparameters}
\begin{figure}[!htp]
  \centering
  \includegraphics[width=0.75\textwidth]{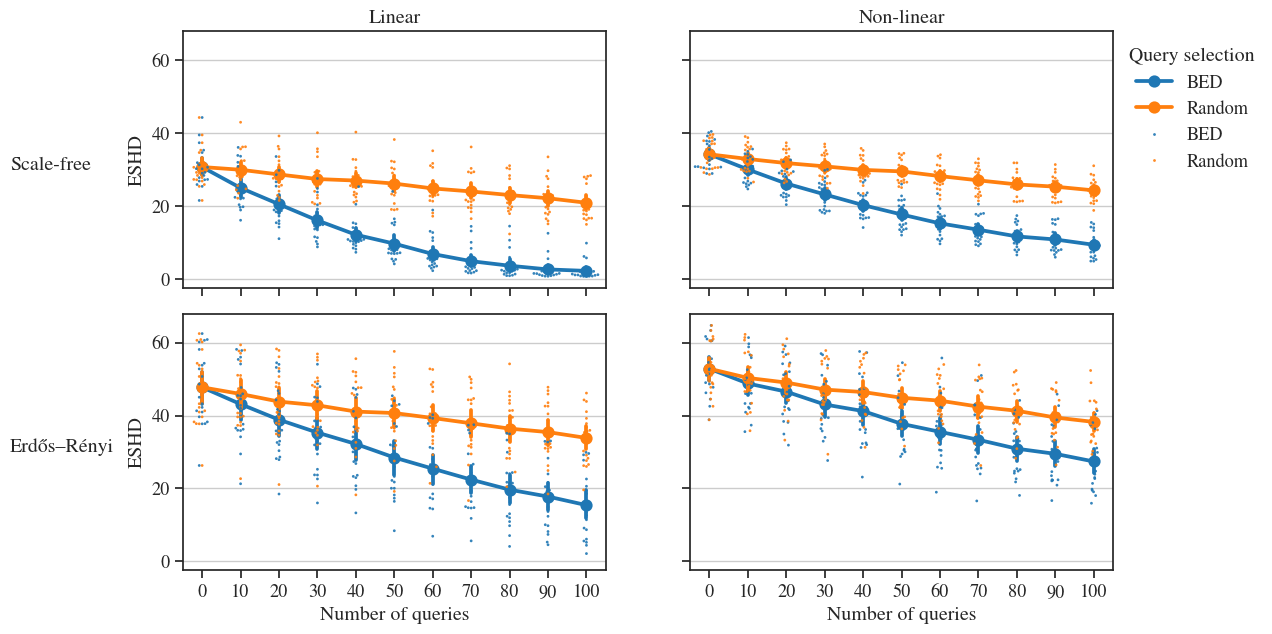}
  \caption{Mean and 95\% (bootstrapped) confidence interval of ESHD for VaMSL with more informative prior using hyperparameters $\alpha_0 = \beta_0=50$. Top panels correspond to scale-free graphs and bottom panels to Erd\H{o}s-R\'enyi. Left panels show linear parametrizations. Right panels show non-linear parametrizations.}
  \label{fig:prior_hyperparameter_ablation_higher}
\end{figure}

Below we demonstrate the effect of varying the informativeness of our proposed (latent) graph prior. The informativeness is controlled by the hyperparameters $\alpha_0, \beta_0$ which, while these could vary edge-wise if the expert had opinions of different strength for given edges, we take to be equal for all edges and $\alpha_0 = \beta_0$. As can be seen from \ref{fig:prior_hyperparameter_ablation_higher}, the results of the inference are more dramatically impacted when using a higher informativeness. The experimental setup is otherwise identical to that used in \Cref{sec:results_homogeneous}. 

\begin{figure}[H]
    \centering
  \includegraphics[width=0.75\textwidth]{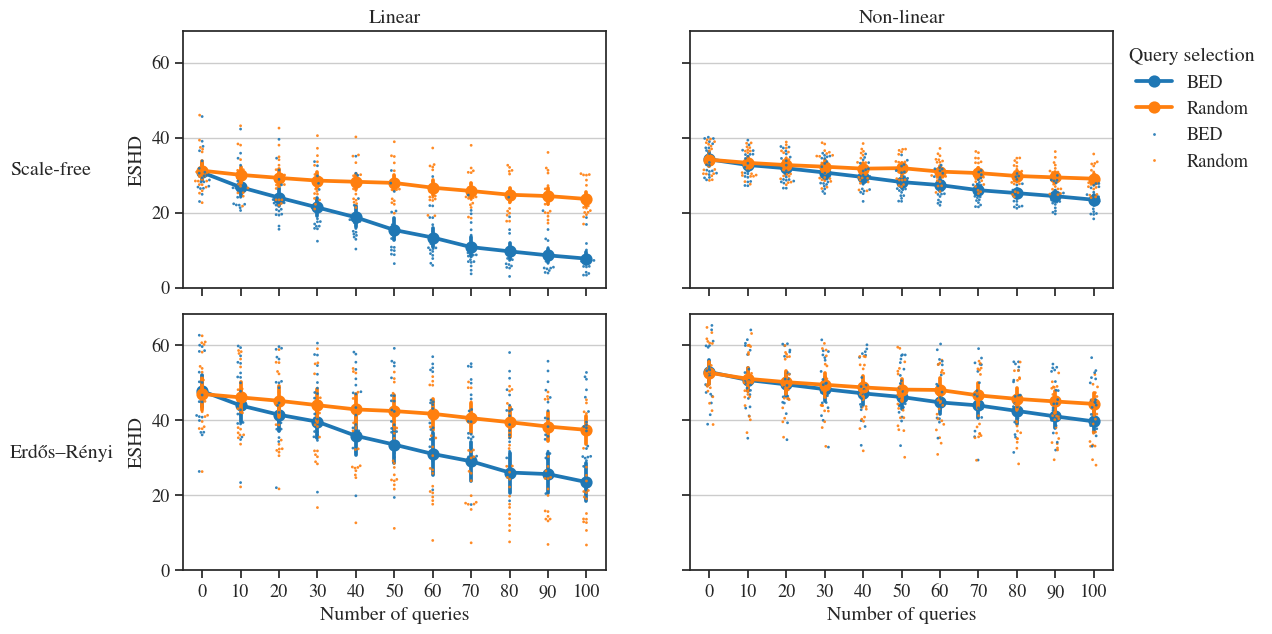}
    \caption{Mean and 95\% (bootstrapped) confidence interval of ESHD for VaMSL with less informative prior using hyperparameters $\alpha_0 = \beta_0=2$. Top panels correspond to scale-free graphs and bottom panels to Erd\H{o}s-R\'enyi. Left panels show linear parametrizations. Right panels show non-linear parametrizations.}
    \label{fig:prior_hyperparameter_ablation_lower}
\end{figure}

\subsection{Ablation Studies on \texorpdfstring{$\alpha_0$}{alpha 0} and Reliability}
\begin{figure}[!htp]
  \centering
  \subfloat{\includegraphics[width=0.75\textwidth]{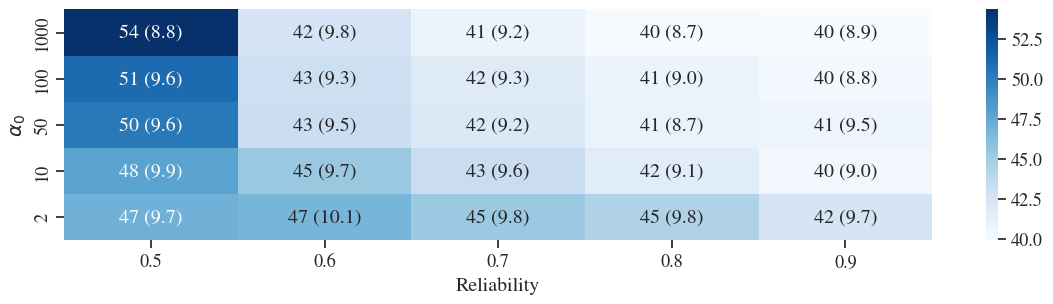}}
  \hfill
  \subfloat{\includegraphics[width=0.75\textwidth]{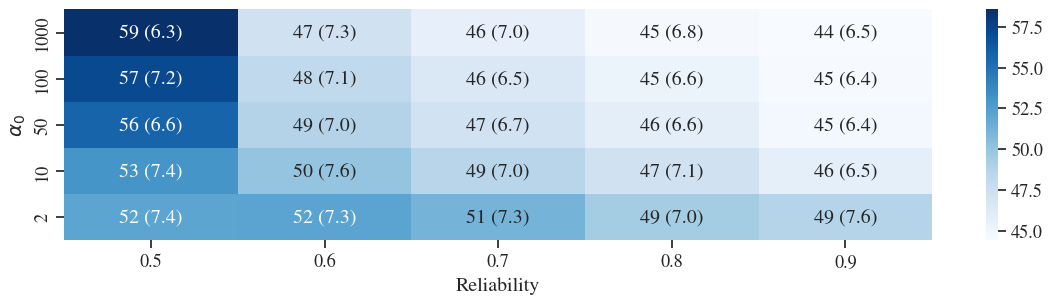}}
  \caption{Average ESHD (SD) for varying levels of $\alpha_0$ and expert reliability in ER graphs, over 20 independent runs. For \textbf{linear} (\textit{top}) and \textbf{non-linear} graphs (\textit{bottom}).}
  \label{fig:ER-alpha_reliability}
\end{figure}

\begin{figure}[!htp]
  \centering
  \subfloat{\includegraphics[width=0.75\textwidth]{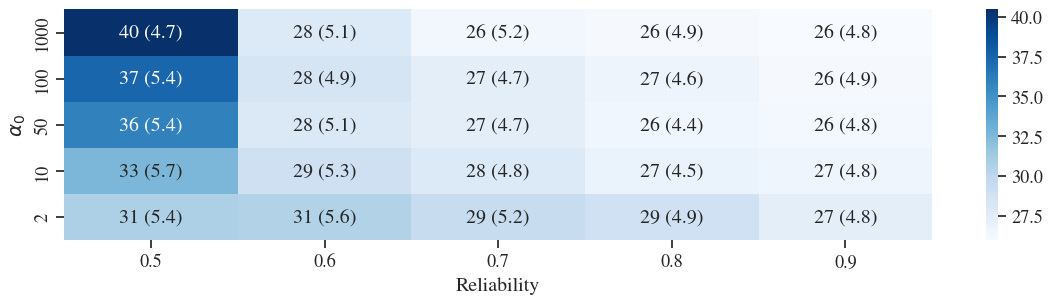}}
  \hfill
  \subfloat{\includegraphics[width=0.75\textwidth]{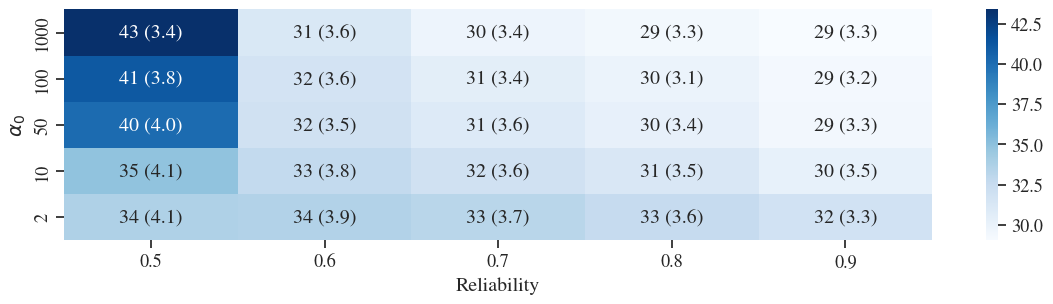}}
  \caption{Average ESHD (SD) for varying levels of $\alpha_0$ and expert reliability in SF graphs, over 20 independent runs. For \textbf{linear} (\textit{top}) and \textbf{non-linear} graphs (\textit{bottom}).}
  \label{fig:SF-alpha_reliability}
\end{figure}

Figures~\ref{fig:ER-alpha_reliability} and~\ref{fig:SF-alpha_reliability} display the performance (measured by ESHD) of different degrees of informativeness, as encoded by $\alpha_0$, fare against different levels of reliability. When the expert is highly reliable, the informativeness of the prior does not seem to have a significant effect, and we obtain similar performance through out. On the other hand, for lower levels of expert-reliability, the informativeness of the parameter is key to improve the performance. Other than the expert reliability and $\alpha_0$, the experimental settings are identical to those in in Section~\ref{sec:application}.

The decreased structure learning when the reliability is set to $0.5$ and the informativeness is increased is due to the implementation of the expert oracle (as explained in \Cref{app:expert-simulation}). When the oracle's mean is set to $0.5$ the distribution for the responses will be equally probable to give slightly incorrect answers as correct ones, leading to the increased informativeness of the prior emphasizing incorrect beliefs about the ground truth graph.  

\subsection{Ablation Study of the Expert Model}\label{app:expert-model}
In Tables~\ref{table:nts_notears_ER}~and~\ref{table:nts_notears_SF} we compare our proposed expert model using soft-constraints against NTS-NOTEARS \citep{sun_nts-notears_2023}, which needs binary responses. This showcases how binarized responses from an expert fail to correctly capture the nuances that continuous responses are able to capture. The approach we follow is to map the expert's responses, originally in $(0, 1)$, to $\{0,1\}$ by using a rounding function. In principle, this should lead to an advantage to NTS-NOTEARS as the given responses are correct and hard constraints have a bigger impact than soft constraints. However, as displayed below, in this setting VaMSL outperforms NTS-NOTEARS irrespective of the number of queries.

\begin{table}[!ht]
    \centering
    \caption{Average (SD) of (E)SHD for \textbf{nonlinear} Gaussian ER BNs with 20 variables across 10 replications with randomly selected queries. NTS-NOTEARS was run with the Additive Noise Model (ANM) and default settings. All methods using a single component.}
    \begin{tabular}{ l | r r r r r r}
    \toprule
         Method & 0 queries & 20 queries & 40 queries & 60 queries & 80 queries & 100 queries \\ \midrule
        VaMSL + expert & 51 (8.6) & 47 (7.6) & 45 (7.7) & 44 (6.8) & 41 (7.2) & 38 (5.9) \\ 
        NTS-NOTEARS & 59 (6.2) & 57 (4.4) & 54 (7.0) & 54 (5.7) & 50 (5.9) & 51 (9.5) \\ \bottomrule
    \end{tabular}
    \label{table:nts_notears_ER}
\end{table}

\begin{table}[!ht]
    \centering
    \caption{Means and standard deviations of (E)SHD for \textbf{nonlinear} Gaussian SF BNs with 20 variables  across 10 replications with randomly selected queries. NTS-NOTEARS was run with the Additive Noise Model (ANM) and default settings. All methods using a single component.}
    \begin{tabular}{ l | r r r r r r}
    \toprule
        Method & 0 queries & 20 queries & 40 queries & 60 queries & 80 queries & 100 queries \\ \midrule
        VaMSL + expert & 34 (4.2) & 32 (3.5) & 30 (3.4) & 29 (3.2) & 27 (3.4) & 25 (2.9) \\ 
        NTS-NOTEARS & 83 (7.2) & 77 (6.1) & 73 (7.6) & 68 (5.4) & 64 (8.5) & 55 (6.8)  \\ \bottomrule
    \end{tabular}
    \label{table:nts_notears_SF}
\end{table}

\subsection{Computational Scalability of VaMSL to Higher Dimensions}\label{app:scalability}
Tables~\ref{table:scalability_linear_ER}-\ref{table:scalability_nonlinear_SF} show the computational scalability of VaMSL on two components to higher number of variables by displaying the running times for the different combinations of nonlinear-linear and scale-free and Erd\H{o}s-R\'enyi graphs. Overall, the results match the scale of running times stated in Appendix~\ref{sec:vamsl-algorithm}.

\begin{table}[ht!]
    \centering
    \caption{Means and standard deviations of computation times (sec.) for \textbf{linear} Gaussian ER mixtures with two components across 10 replications.}
    \begin{tabular}{c | c c c}
    \toprule
        N particles & 10 Variables & 20 Variables & 50 Variables \\ \midrule
        10 & 71 (2.1) & 106 (2.2) & 444 (328.4) \\
        30 & 146 (2.8) & 265 (4.3) & 928 (6.5) \\ \bottomrule
    \end{tabular}
    \label{table:scalability_linear_ER}
\end{table}

\begin{table}[ht!]
    \centering
    \caption{Means and standard deviations of computation times (sec.) for \textbf{nonlinear} Gaussian ER mixtures with two components across 10 replications.}
    \begin{tabular}{c | c c c}
    \toprule
        N particles & 10 Variables & 20 Variables & 50 Variables \\ \midrule
        10 & 402 (147.5) & 1113 (1189.4) & 2960 (2061.3) \\
        30 & 775 (269.3) & 2042 (1210.3) & 10269 (6277.1) \\ \bottomrule
    \end{tabular}
    \label{table:scalability_nonlinear_ER}
\end{table}

\begin{table}[!ht]
    \centering
    \caption{Means and standard deviations of computation times (sec.) for \textbf{linear} Gaussian \textbf{SF} mixtures with two components across 10 replications.}
    \begin{tabular}{c | c c c}
    \toprule
        N particles & 10 Variables & 20 Variables & 50 Variables \\ \midrule
        10 & 70 (1.8) & 107 (1.4) & 338 (4.0) \\
        30 & 173 (4.4) & 306 (6.1) & 1779 (428.4) \\ \bottomrule
    \end{tabular}
    
    \label{table:scalability_linear_SF}
\end{table}

\begin{table}[!ht]
    \centering
    \caption{Means and standard deviations of computation times (sec.) for \textbf{nonlinear} Gaussian \textbf{SF} mixtures with two components across 10 replications.}
    \begin{tabular}{c | c c c}
    \toprule
        N particles & 10 Variables & 20 Variables & 50 Variables \\ \midrule
        10 & 316 (93.1) & 761 (534.4) & 2219 (776.1) \\ 
        30 & 813 (309.6) & 2346 (1546.8) & 7100 (3576.2) \\ \bottomrule
    \end{tabular}
    \label{table:scalability_nonlinear_SF}
\end{table}

\subsection{Behavior of VaMSL against Misspecification of the Number of Components}\label{app:component-misspecification}
In Tables~\ref{table:component_misspecification_linear_ER}-\ref{table:component_misspecification_nonlinear_SF} we show how VaMSL behaves when the number of components is misspecified. We consider two possible sources of model misspecification. First, there are two components in the true data generating process and the model incorrectly assumes there are three mixture components. Second, there are three components in the data generating process, and the model assumes there are two mixture components. In each experiment, the components in the models have an equal share of a total of 60 particles. For these experiments, we report the weighted ESHD, defined as:
\begin{equation*}
    \textrm{wESHD} = \sum_n\sum_i\sum_k c_{ni} q(c_{nk}=1)\textrm{ESHD}(p(G_k\mid\mathcal{D}), G^*_i),
\end{equation*}
where $G^*_i$ refers to the ground truth graph of the $i$\textsuperscript{th} DGP and $c_{ni}=1$ when the observation $\mathbf{x}_n$ was generated from the DGP associated with $G^*_i$, while otherwise $c_{ni}=0$.

The results show how, although choosing to few components will adversely affect the inference, when there are more components than DGPs the inference does not suffer. This is due to the mixture model either degenerating the responsibilities of the additional component, rendering the mixture correctly specified, or using it to infer the same graph as one of the other components thereby not negatively impacting the inference.

\begin{table}[!ht]
    \centering
    \caption{Means and standard deviations of weighted ESHD for \textbf{linear} Gaussian ER mixtures across 10 replications. All mixture components have the same probability (1/number of components).}
    \begin{tabular}{c | c c c}
    \toprule
        Number of DGPs & 1 Component & 2 Components & 3 Components \\ \midrule
        2 & 72 (3.8) & 44 (6.3) & 45 (8.5) \\
        3 & 71 (2.6) & 68 (7.0) & 55 (10.7) \\ \bottomrule
    \end{tabular}
    \label{table:component_misspecification_linear_ER}
\end{table}

\begin{table}[!ht]
    \centering
    \caption{Means and standard deviations of weighted ESHD for \textbf{nonlinear} Gaussian ER mixtures across 10 replications. All mixture components have the same probability (1/number of components).}
    \begin{tabular}{c|c c c}
    \toprule
        Number of DGPs & 1 Component & 2 Components & 3 Components \\ \midrule
        2 & 77 (7.5) & 62 (11.3) & 57 (10.1) \\ 
        3 & 84 (5.1) & 72 (11.2) & 60 (11.8) \\ \bottomrule
    \end{tabular}
    \label{table:component_misspecification_nonlinear_ER}
\end{table}

\begin{table}[!ht]
    \centering
    \caption{Means and standard deviations of weighted ESHD for \textbf{linear} Gaussian SF mixtures across 10 replications. All mixture components have the same probability (1/number of components).}
    \begin{tabular}{c|c c c}
    \toprule
        Number of DGPs & 1 Component & 2 Components & 3 Components \\ \midrule
        2 & 58 (3.2) & 45 (10.8) & 42 (9.4) \\
        3 & 64 (2.2) & 56 (4.2) & 47 (6.7) \\ \bottomrule
    \end{tabular}
    \label{table:component_misspecification_linear_SF}
\end{table}

\begin{table}[!ht]
    \centering
    \caption{Means and standard deviations of weighted ESHD for \textbf{nonlinear} Gaussian SF mixtures across 10 replications. All mixture components have the same probability (1/number of components).}
    \begin{tabular}{c|c c c}
    \toprule
        Number of DGPs & 1 Component & 2 Components & 3 Components \\ \midrule
        2 & 60 (6.7) & 44 (12.6) & 41 (9.2) \\
        3 & 78 (10.4) & 62 (12.6) & 45 (9.1) \\ \bottomrule
    \end{tabular}
    \label{table:component_misspecification_nonlinear_SF}
\end{table}

\end{appendices}
\end{document}